\documentclass[letterpaper]{article} 
\usepackage{aaai23}  
\usepackage{times}  
\usepackage{helvet}  
\usepackage{courier}  
\usepackage[hyphens]{url}  
\usepackage{graphicx} 
\usepackage{natbib}  
\usepackage{caption} 
\usepackage{algorithm}
\usepackage{algorithmic}

\usepackage{newfloat}
\usepackage{listings}
\DeclareCaptionStyle{ruled}{labelfont=normalfont,labelsep=colon,strut=off} 
\floatstyle{ruled}
\newfloat{listing}{tb}{lst}{}
\floatname{listing}{Listing}
\usepackage{amsfonts}
\usepackage{amsmath}
\usepackage{bm}
\usepackage{mathtools}
\usepackage{amsthm}
\usepackage{subcaption}
\usepackage{booktabs}
\usepackage{amssymb}
\usepackage{multirow}
\usepackage{color}

\newtheorem{lemma}{Lemma}
\newtheorem{theorem}{Theorem}

\title{On the Limit of Explaining Black-box Temporal Graph Neural Networks}
\author {
    Minh N. Vu,\textsuperscript{\rm 1}
    My T. Thai \textsuperscript{\rm 1}
}
\affiliations {
    \textsuperscript{\rm 1} \textit{University of Florida},
Gainesville, Florida, USA\\
    minhvu@ufl.edu, mythai@cise.ufl.edu
}

\begin{document}

\maketitle


\begin{figure*}[ht]
		\centering
         \begin{subfigure}{.36\linewidth}
          \centering
          \includegraphics[height=2.4cm]{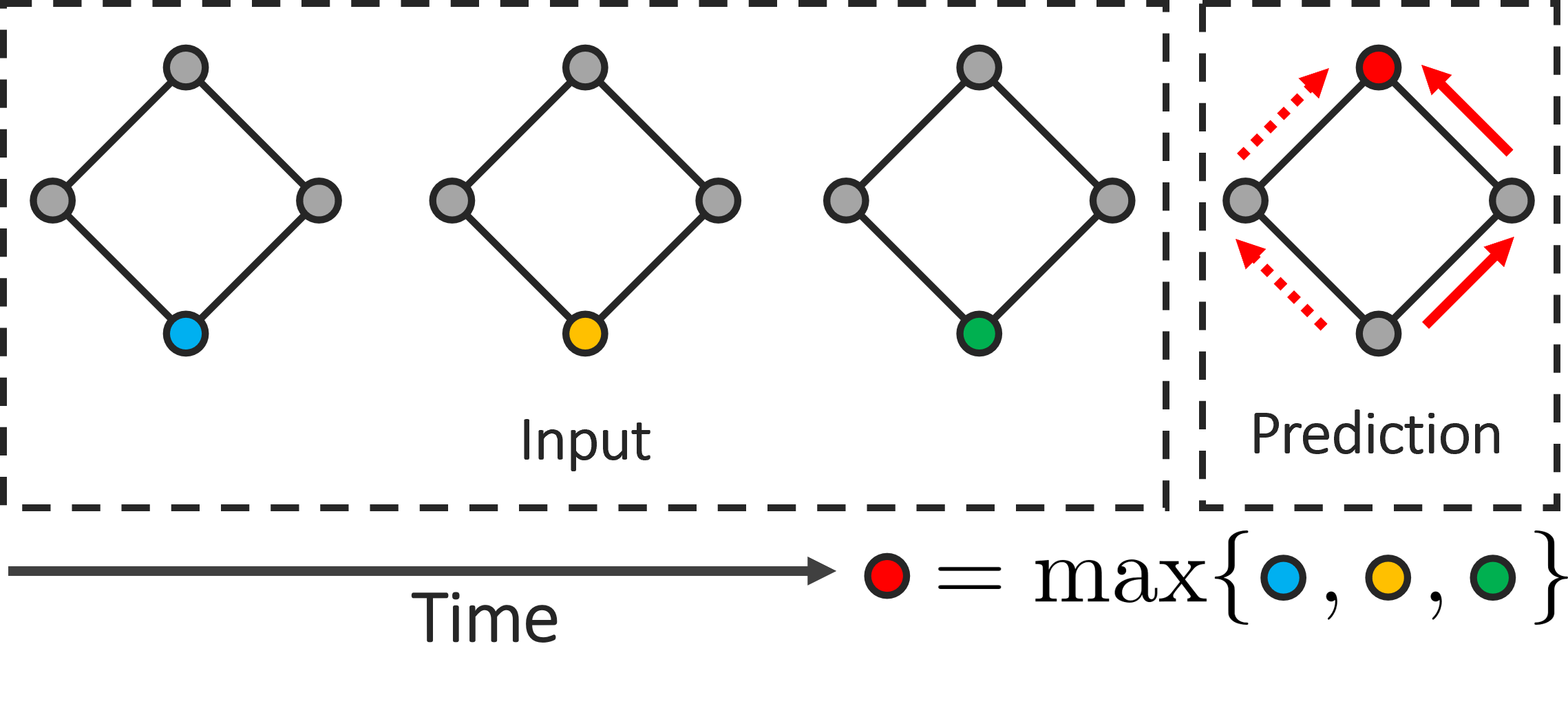}
          \caption{Multi-path aggregation task.}
                \label{fig:task_node_top}
        \end{subfigure}
        \begin{subfigure}{.31\linewidth}
          \centering
          \includegraphics[height=2.4cm]{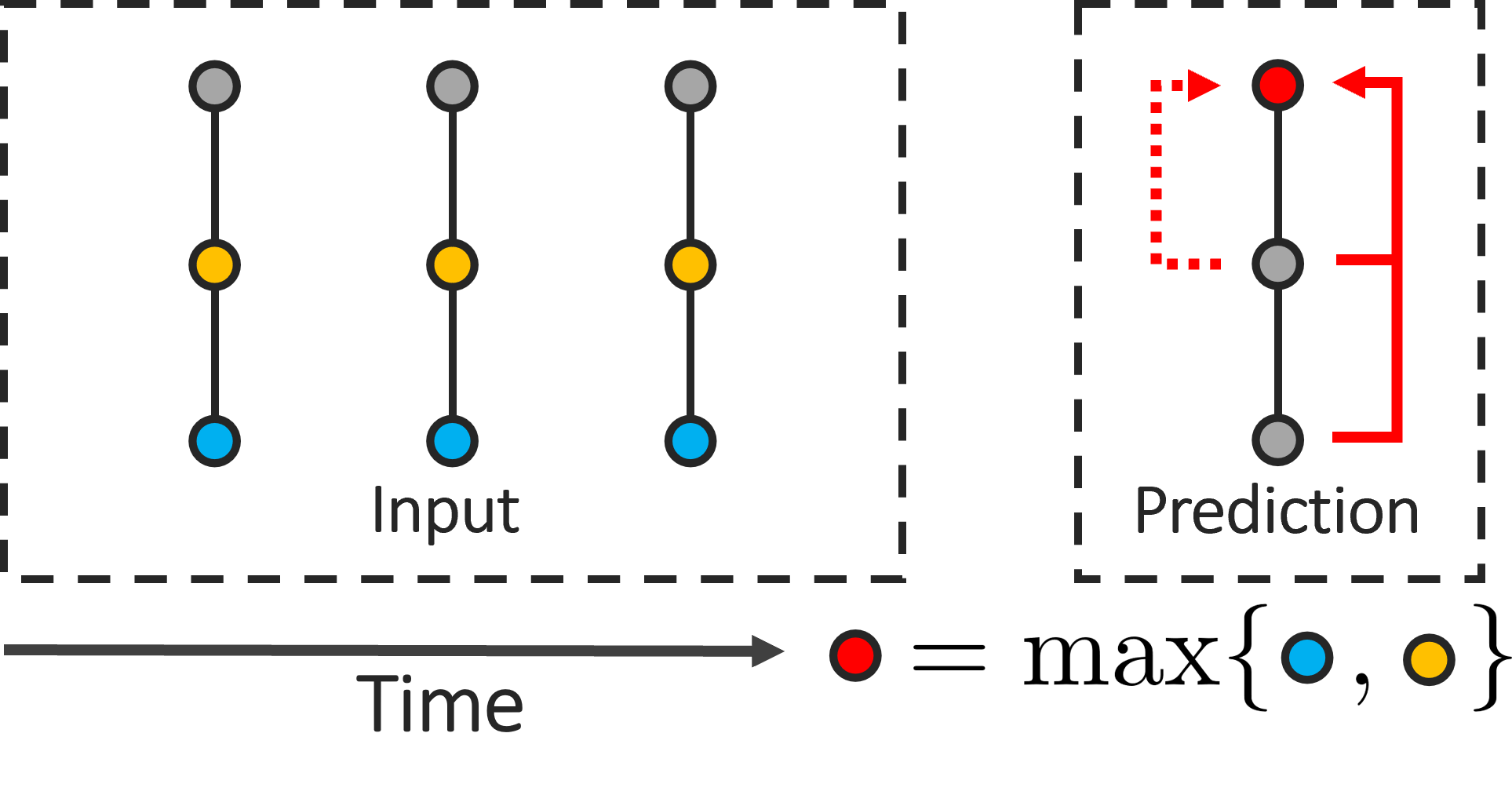}
          \caption{Multi-node aggregation task.}
                \label{fig:task_edge_top}
        \end{subfigure}
        \begin{subfigure}{.31\linewidth}
          \centering
          \includegraphics[height=2.4cm]{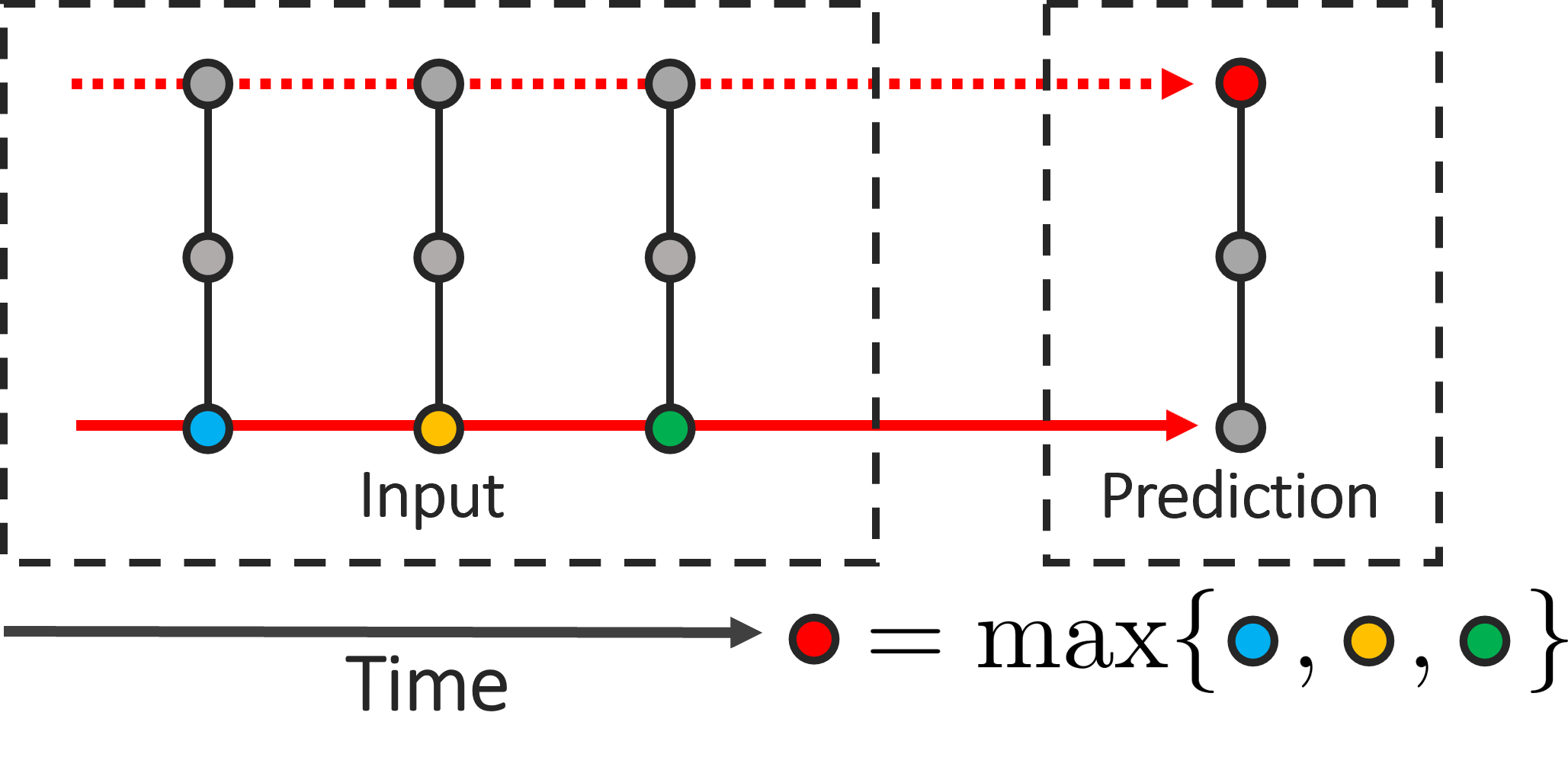}
          \caption{Temporal aggregation task}
                \label{fig:task_node_edge_top}
        \end{subfigure}%
        \caption{Tasks for the class of (a) Node-perturbation, (b) Edge-perturbation and (c) Node-and-Edge-perturbation explanation methods. The dash arrows and the dotted arrows show different internal computations that the model can carry out. Unidentifiable Proofs show that explanation methods cannot differentiate the computations; thus, cannot identify those internal dynamic.}
        \label{fig:unidentifiable_task}
\end{figure*}

\begin{abstract}
Temporal Graph Neural Network (TGNN) has been receiving a lot of attention recently due to its capability in modeling time-evolving graph-related tasks. Similar to Graph Neural Networks, it is also non-trivial to interpret predictions made by a TGNN due to its black-box nature. 
A major approach tackling this problems in GNNs is by analyzing the model' responses on some perturbations of the model's inputs, called perturbation-based explanation methods. While these methods are convenient and flexible since they do not need internal access to the model, does this lack of internal access prevent them from revealing some important information of the predictions? Motivated by that question, this work studies the limit of some classes of perturbation-based explanation methods. Particularly, by constructing some specific instances of TGNNs, we show (i) node-perturbation cannot reliably identify the paths carrying out the prediction, (ii) edge-perturbation is not reliable in determining all nodes contributing to the prediction and (iii) perturbing both nodes and edges does not reliably help us identify the graph's components carrying out the temporal aggregation in TGNNs. 
\end{abstract}

\section{Introduction}

Graph Neural Networks (GNNs) have been achieving successful performance in many practical graph-related problems including social networks, citation networks, knowledge graphs, and biological networks~\cite{You_NIPS2018,Zhang_NIPS2018,Zitnik2018}. Many architectures with nice design and high predictive performance have been introduced in recent years~\cite{Defferrard2016,Kipf2016iclr, GraphSage, velickovic2018graph, xu2018how}.
Along those works, a notable branch of GNNs has been developed with the goal to integrate temporal information into the graph structure, called Temporal Graph Neural Networks (TGNN)~\cite{zhao2019t,wang2020traffic,min2021stgsn}. This variant has shown promising outcomes in domain where the data has strong correlation with time such as transportation and weather forecast.

Since GNNs and TGNNs inherit the black-box nature of neural networks, interpreting their predictions remains a daunting task as no internal information of the models, such as the hidden activation or the gradients, is available. Many explanation methods, called {explainers}, are introduced to explain local predictions of GNNs in this setting~\cite{Ying_NIPS2019, duval2021graphsvx, minhpgm, GraphLIME}. These methods generally rely on the model's responses on some perturbations of the input to solve for good explanations. While the approach has shown many heuristic successes, there is little theoretical result on its performance. 
In particular, is there any information on the model's internal behavior that a given method of perturbation cannot reveal? 
Addressing this question will help us design better explainers for variants of GNNs, such as TGNNs and many other architectures to come. More importantly, analyzing the limit of perturbation-based methods also help identify what information should not be read out from some given explanations.

Motivated by 
the above observations, our work focuses on the limit of perturbation-based explanation methods when applying to TGNN, i.e. what information cannot be revealed by a given class of perturbation. The classes are determined by the input's components that they perturb, which are node-only, edge-only, and node-and-edge. Our analysis follows a proof structure, called \textit{Unidentifiable Proof}, through which the limit of perturbation methods can be formalized and analyzed. For each class of explainers, we identify a training task (Fig.~\ref{fig:unidentifiable_task}) and construct a model such that there is no method in the class can identify certain internal dynamics of the model in generating its prediction. Specifically, given a constant $K$ determined by the model's parameters, we show:
\begin{itemize}
    \item Node-perturbation methods bounded by $K$ cannot identify the path carrying out the message passing. (Fig.~\ref{fig:task_node_top}).
    \item Edge-perturbation methods cannot identify all nodes contributing to a max aggregation. (Fig.~\ref{fig:task_edge_top}).
    \item Node-and-Edge-perturbation methods bounded by $K$ cannot identify which nodes carry out the temporal aggregation. (Fig.~\ref{fig:task_node_edge_top}).
\end{itemize}
In practice, $K$ is the result of training and can be arbitrarily large. Thus, our analysis is applicable to many practical explainers as large perturbations are often weighted lightly due to the notion of locality~\cite{Marco2016}.
Although some of our results are applicable to GNNs, we focus our analysis on TGNNs because there are few works explaining them, which makes the study more meaningful for the future research. Another reason is, as TGNNs add a temporal dimension to GNNs, it also introduces a temporal dimension to the explaining problem. We find studying this temporal aspect is novel and interesting by itself. 


The outline of this manuscript is as follows. Sect.~\ref{sect:related_work} and Sect.~\ref{prelim} discuss the related works and preliminaries, respectively. 
Our Unidentifiable Proof and related notions are introduced in Sect.~\ref{sect:unidentifiable}. Sects.~\ref{sect:node},~\ref{sect:edge} and \ref{sect:node_edge} follow our Unidentifiable Proof and formally demonstrate the type of information that Node-perturbation, Edge-perturbation and Node-and-Edge-perturbation cannot identify.
Sect.~\ref{sect:exp} provides some synthetic and real-world experiments showing the impact of perturbation schemes on the explaining tasks.
Sect.~\ref{sect:discussion} concludes the paper with several interesting discussions on the implications of our results and their practical aspects. 

\section{Related Works} \label{sect:related_work}


To our knowledge, there is currently no work theoretically study the limitation of explanations for GNNs or TGNNs. Even though extensive experiments evaluating explanation methods have been conducted~\cite{NEURIPS2020_417fbbf2,Amara_2022}, there exist many pitfalls and challenges in those evaluations as there is a mismatch between the ground-truth and the GNN~\cite{Faber2021WhenCT}. Furthermore, with the increasing number of dataset, model architectures and explanation methods, conducting comprehensive evaluations is becoming much more challenging, especially for black-box methods 
of which the computation complexity is significantly higher than that of white-box methods~\cite{Amara_2022}.

Our work is directly related to black-box perturbation-based explaining methods for GNNs, including GNNExplainer~\cite{Ying_NIPS2019}, {PGExplainer}~\cite{luo2020parameterized}, GraphLIME~\cite{GraphLIME}, PGMExplainer~\cite{minhpgm}, {RelEx}~\cite{Zhang2021RelExAM}, GraphSVX~\cite{duval2021graphsvx} and ZORRO~\cite{funke2021zorro}. The perturbations in those methods can either be determined at the start of the algorithms, or be computed iteratively during some specific optimizations. Table~\ref{table:scope_summary} summaries the perturbation methods used by those explainers and the scope of results in each section of this paper.

\begin{table}[ht]
\centering
\caption{Summary of perturbation methods used by explainers and the scope of our results.}
\label{table:scope_summary}
\scalebox{0.9}{
\begin{tabular}{@{}cccllc@{}}
\toprule
\textbf{} & Node                 & Edge                 & \multicolumn{1}{c}{Sect.\ref{sect:node}} & \multicolumn{1}{c}{Sect.\ref{sect:edge}} & Sect.\ref{sect:node_edge} \\ \midrule
GNNExplainer     & *                    & *                    &                                                                 &                                                                 & *                                                 \\
PGExplainer      & \multicolumn{1}{l}{} & *                    &                                                                 & \multicolumn{1}{c}{*}                                           & *                                                 \\
GraphLIME     & *                    & \multicolumn{1}{l}{} & \multicolumn{1}{c}{*}                                           &                                                                 & *                                                 \\
PGMExplainer     & *                    & \multicolumn{1}{l}{} & \multicolumn{1}{c}{*}                                           &                                                                 & *                                                 \\
RelEx     & \multicolumn{1}{l}{} & *                    &                                                                 & \multicolumn{1}{c}{*}                                           & *                                                 \\
GraphSVX      & *                    &                     &   \multicolumn{1}{c}{*}                                                                &                                                                 & *                                                 \\
ZORRO     & *                    & \multicolumn{1}{l}{} & \multicolumn{1}{c}{*}                                           &                                                                 & *                                                 \\ \bottomrule
\end{tabular}
}
\end{table}







\section{Preliminaries} \label{prelim}
We now provide some preliminaries and notations that are commonly used in the researches of GNNs and the explaining problem. We also briefly introduce Dynamic Bayesian Networks, which we use in our Unidentifiable Proofs.

\textbf{Notation.} For all models studied in this work, their inputs are defined on a graph $G = (V,E)$, where $V$ and $E$ are the set of nodes and edges, respectively. For TGNNs, the inputs are a sequence of feature vectors $ \bm X_{t_s, t_e} \coloneqq [X^{(t_s)}, \cdots, X^{(t_e)}]$ and an adjacency matrix $ A \in \mathcal{A} \coloneqq \{0,1\}^{|V|\times |V|}$. Here, $t_s$, $t_e$ and $X^{(t)} \in \mathbb{R}^{|V|\times F}$ denote the starting time, the ending time and the feature matrices of the input sequence. The model is referred by its forwarding function $\Phi: \mathcal{X} \times \mathcal{A} \rightarrow \mathcal{Y}$, where $\mathcal{X}$ and $\mathcal{Y}$ are the space of feature matrix-sequence $ \bm X_{t_s, t_e} $ and the model's output.

\textbf{Graph Neural Networks.} We use the general formulation of GNNs based on the message passing mechanism~\cite{GraphSage}, which involves 3 computations: propagation, aggregation and update:
\begin{align*}
    &m_{ij}^{(l)} = \textup{MSG} \left(  h_i^{(l-1)},  h_j^{(l-1)} \right), \\
    &a_i^{(l)} = \textup{AGG} \left( \left \{ m_{ji}^{(l)} \right\}_{j \in \mathcal{N}_i} \right),  
    h_i^{(l)} = \textup{UPD} \left( a_i^{(l)}, h_i^{(l-1)} \right)
\end{align*}
Here $m_{ij}$ is the message from node $i$ to node $j$, $h_i^{(l)}$ is the hidden representations of node $i$ at layer $l$ and $\mathcal{N}_i$ is node $i$'s neighbors.
The final representation at the last layer $L$, $ h_i^{(L)}$, is commonly used to generate a prediction, i.e. $ Y = \textup{READOUT} ( h_i^{(L)} )$. Typically, the MSG, UPD, and READOUT functions consist of trainable weights and biases followed by an activation. The AGG is commonly chosen as a max, mean, or concatenation aggregation.

\textbf{Temporal Graph Neural Networks.} The forwarding function $\Phi: \mathcal{X} \times \mathcal{A} \rightarrow \mathcal{Y}$ of a TGNN can be reformulated based on its sequential implementation~\cite{zhao2019t}:
\begin{align}
    H^{(t_s)} &= \bar{\Phi} (X^{(t_s)},  A)  \nonumber \\
    H^{(t)} &= \bar{\Phi} (H^{(t-1)}, X^{(t)},  A), \ t= t_s+1, ... t_e   \label{eq:tgnnlast}
\end{align}
where $\bar{\Phi}$ is the forwarding function of a GNN and $ H^{(t)}$ is the temporal messages. $T \coloneqq t_e - t_s + 1$ is the input's length. 

The {\em base} GNN $\bar{\Phi}$ typically consists of some graph layers followed by a readout. The output $Y$ is computed either by applying a readout on the temporal message at the last layer $H^{(t_e)}$ or from the node's final hidden features. In this manuscript, capital letters, e.g. $X,Y$ and $H$, refer to external signals of the GNN blocks, while small letters, e.g. $m, a$ and $h$, are for internal signals.

\textbf{The Class of Explainers.} 
This work studies black-box explainers of GNNs and TGNNs based on the type of perturbations that the explainers use:
    \begin{itemize}
        \item Node-perturbation class $\mathcal{G}_v$: the explainer can perturb the entries of the feature matrices in $\bm X_{t_s, t_e}$.
        \item Edge-perturbation class $\mathcal{G}_e$: the explainer can remove some edges from the input, i.e. zeroing out some entries of the input adjacency matrix $ A$.
        \item Node-and-Edge-perturbation class $\mathcal{G}_a$:  the explainer can perturb both the feature matrices in $\bm X_{t_s, t_e}$ and remove some edges in the input adjacency matrix $ A$.
    \end{itemize}

\textbf{Dynamic Bayesian Networks (DBNs).} The usage of DBNs in this work is to model internal computations of TGNNs so that theoretical analysis can be conducted. A DBN~\cite{Dubois1992} can be considered as an extension of Bayesian networks (BNs)~\cite{PEARL198877} to model temporal dependency of systems' variables. Temporal information is integrated via edges between adjacent time steps. More details of BNs and DBNs are in Appx.~\ref{appendix:prelim_bn}.

\section{Unidentifiable Proof for Neural Networks} \label{sect:unidentifiable}

Given a black-box model $\Phi$ and a class of explanation methods, the Unidentifiable Proof formalizes the idea that certain information of $\Phi$ cannot be identified by explainers belong to that class. However, before introducing the Unidentifiable Proof, we need to address an issue hindering this study, which is about the ground-truth explanation. It is important to formalize this notion because, without it, the study of explanation methods cannot be rigorous. The first two subsections discussing about the \textit{interpretable domain} and the \textit{Transparent Model} serve that purpose. Intuitively, the interpretable domain is the domain of all available explanations and the Transparent Model is a domain's member that can faithfully capture the model. Given that, the latter part of this section describes the Unidentifiable Proof. 

The general idea of the Unidentifiable Proof is by construction: it constructs two instances of the model whose Transparent Models are different but their information extracted by a class of explainers are exactly the same. Given such constructions, we can conclude that no explainer of that class can distinguish the two instances. Furthermore, as we will show, they also cannot identify any model's dynamics that can be used to differentiate the Transparent Models. This gives us formal notions of unidentifiable information.




\subsection{The Interpretable Domain}
Given a black-box model $\Phi$ and an input $\bm X$, the explainers solve for an interpretable representation of the prediction $\Phi (\bm X)$, denoted as $g (\Phi (\bm X))$. For the sake of explaining, $g (\Phi (\bm X))$ should be intuitive and interpretable; therefore, we denote the domain of $g (\Phi (\bm X))$ by the \textit{interpretable domain}. For example, the interpretable domains for GNNs have been chosen to be a set of scores on some nodes/edges' features, the set of linear functions and the set of probabilistic models on the input's nodes~\cite{GraphLIME,duval2021graphsvx,minhpgm}.  Intuitively, a good interpretable domain should balance between its representative power and interpretability. In this work, we consider it to be the set of DBNs. We will describe in more details how DBNs can help explain TGNNs in the next subsection.

\subsection{The Transparent Model}

Given an interpretable domain and a black-box model $\Phi$, there is no guarantee that there exists an interpretable representation correctly explains $\Phi$. 
Nevertheless, in some specific contexts, an interpretable representation that can fully capture the black-box model exists.
Particularly, the work~\cite{advlime:aies20} embeds a linear function inside a black-box model, which means that linear function can faithfully capture that black-box. This implies, for a given interpretable domain and for some $\Phi$, explanation that can fully capture $\Phi$ exists. We denote it by \textit{the Transparent Model} $\mathcal{I}$. In some cases, the Transparent Model only exists for some inputs of a subset $\mathcal{S} \subseteq \mathcal{X}$. We write the Transparent Model in those cases as $\mathcal{I} (\Phi (\bm X)), \bm X \in \mathcal{S}$.
Furthermore, we call the assumption $\mathcal{I}(\Phi (\bm X))$ exists  the \textit{Existence assumption}.

\begin{figure}[ht]
\centering
  \includegraphics[width=.9\linewidth]{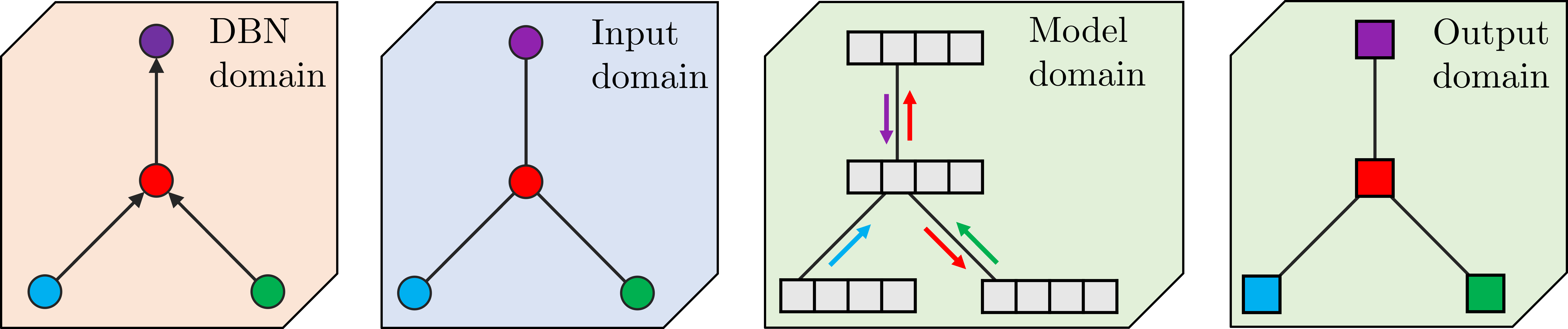}
\caption{The association among variables of the explanation DBN, the input nodes and the messages in the TGNN: Components of the same color are associated with each other.}
\label{fig:interpretable_domain}
\end{figure}

As we consider the interpretable domain to be a set of DBNs, it is important to discuss the Transparent Model $\mathcal{I}(\Phi)$ in term of DBNs. For all our Unidentifiable Proofs, the target of explanation will be the prediction on an input node. The explanation will be in the form of a DBN $\mathcal{B}$, whose variables are associated with the corresponding nodes in the input graph. As each node of the input graph is physically associated with a distinct set of neurons in the graph layers of the TGNN, we then associate each variable of $\mathcal{B}$ to the sending messages of the neurons corresponding to that node in the TGNN. Note that the sending messages from a node consist not only internal messages among neurons but also the temporal messages and the output messages. These associations allow us to capture the dynamics of the TGNN via DBN. Fig.~\ref{fig:interpretable_domain} provides an illustration of these associations. 

We say a DBN $\mathcal{B}$ is the Transparent Model of a TGNN $\Phi$ if (i) all independence claims of $\mathcal{B}$ about its variables are consistent with the sending messages of the corresponding neurons in the TGNN and (ii)  $\mathcal{B}$ is minimal. The condition (i) is obvious since we do not want wrong claims in the explanation. Condition (ii) enforces the explanation DBN to be as informative as possible, i.e. it should remove edges if the variables are independent. 



\subsection{The Unidentifiable Proof}

\begin{figure}[ht]
\centering
  \includegraphics[width=.75\linewidth]{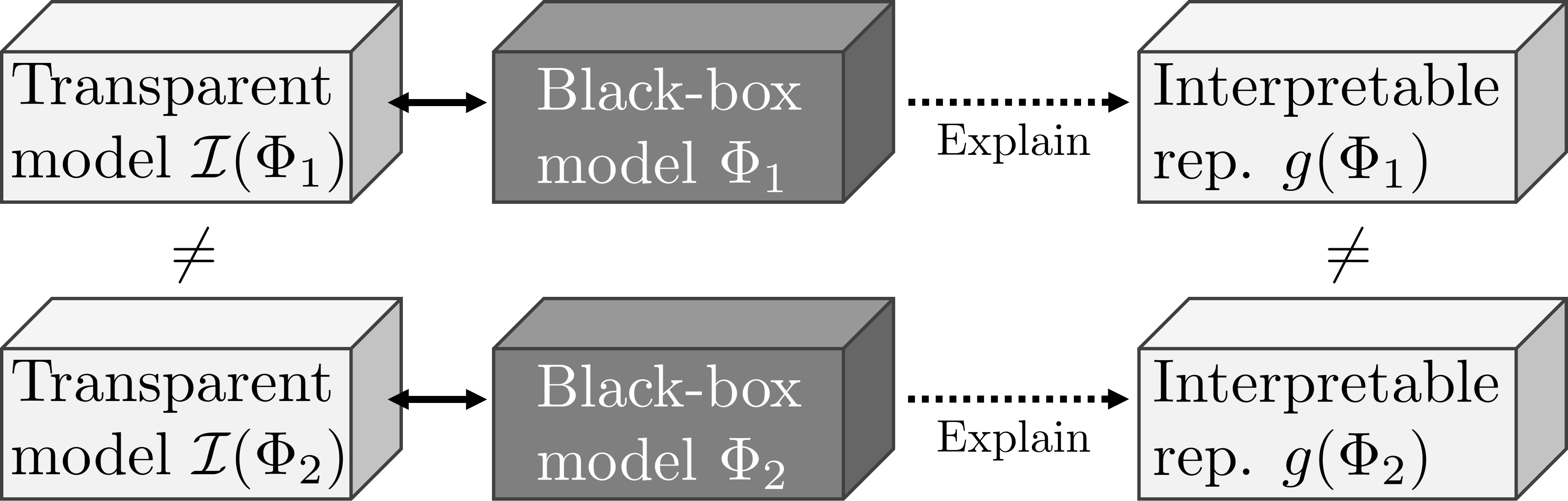}
\caption{Necessary condition for explanations under the Existence assumption.}
\label{fig:proof}
\end{figure}

Under the Existence assumption, i.e. $\mathcal{I}(\Phi (\bm X))$ exists, a good explanation method $g$ is expected to return $g (\Phi (\bm X))$ to be equal or similar to $\mathcal{I}(\Phi (\bm X))$. This provides us a necessary condition to theoretically analyze the limits of explanation methods: given two models $\Phi_1$ and $\Phi_2$ with distinctively different transparent models $\mathcal{I}(\Phi_1)$ and $\mathcal{I}(\Phi_2)$, a good explainer must return an explanation $g(\Phi_1 (\bm X))$ distinctively different from an explanation $g(\Phi_2 (\bm X))$. This necessary condition is demonstrated in Fig.~\ref{fig:proof}.

The above necessary condition can be formalized as follows. Given two neural networks $\Phi_1$ and $\Phi_2$, the Unidentifiable Proof holds if their Transparent Models exist and:
\begin{align}
    &g(\Phi_1(\bm X)) = g(\Phi_2(\bm X)),  \ \forall g \in \mathcal{G}, \forall \bm X \in \mathcal{S} \subseteq \mathcal{X} 
    \label{cond:same}
    \\
     & \exists \bm X \in \mathcal{S} \subseteq \mathcal{X} \textup{ s.t }  \mathcal{I}(\Phi_1(\bm X)) \neq \mathcal{I}(\Phi_2(\bm X)) \label{cond:diff}
\end{align}
The first condition says the explanations of the two models provided by all explainers of class $\mathcal{G}$ are the same. In general, this condition can be shown by examining the forwarding computations. On the other hand, the second condition points out the existence of some inputs such that their Transparent Models are different. The main challenge in proving this condition is in concretely determining $ \mathcal{I}(\Phi_1(\bm X)) $ and $ \mathcal{I}(\Phi_2(\bm X))$. The two conditions then straight-forwardly imply the explainer cannot learn the Transparent Model of at least one of the two models. More interestingly, as $g$ outputs the same information in explaining both models, any information that can be used to differentiate the two models cannot be inferred from $g$. Therefore, all information differentiating $\mathcal{I}(\Phi_1(\bm X)) $ and $ \mathcal{I}(\Phi_2(\bm X))$ also cannot be inferred from the explainer. This then specifies the unidentifiable information of the class of explaining methods $\mathcal{G}$.

\section{Unidentifiable Proof for Node-perturbation} \label{sect:node}

We now provide the Unidentifiable Proof for the Node-perturbation class $\mathcal{G}_v$. We show that for a simple max computation conducted by the TGNNs, Node-perturbation cannot identify the messages' propagating paths carrying out the predictions in the model. We also elaborate how the result can be applied to GNNs in Sect.~\ref{sect:discussion}.

\textbf{The Training Task.}
 In our construction, the TGNNs operate on a graph of 4 nodes forming a square, with the following adjacency matrix:
 \begin{align*}
  \small{A = 
 \begin{bmatrix}
  0&  1&  0& 1\\ 
  1&  0&  1& 0\\ 
  0&  1&  0& 1\\ 
  1&  0&  1& 0
 \end{bmatrix}}
 \end{align*}
The training task is to recognize the maximum positive inputs of node 3 and return result at node 1:
\begin{align}
 Y_1^{(t)} &= \max \{0 \textup{ and } X_3^{(t')} , 0\leq t' \leq t \} \nonumber\\
 Y_2^{(t)} &= Y_3^{(t)} = Y_4^{(t)}  = 0 \label{eq:task_node}
\end{align}
The model's input and output at each time-step $t$ are both in $\mathbb{R}^{4}$. Fig.~\ref{fig:task_node_top} provides an illustration of this training task.

\textbf{The Models.} We construct two TGNNs,  $\Phi^v_1$ and $\Phi^v_2$, with the same architecture but different parameters. We consider the input's length $T = 2$ for the sake of brevity. Each node $i$ is associated with a hidden feature vector $\bm h_i  = [hr_i, ht_i, hs_i, hz_i, ho^+_i, ho^-_i] \in \mathbb{R}^6$, whose features mean:
\begin{itemize}
    \item $hr_i$: message that node $i$ receives.
    \item $ht_i$: temporal message that node $i$ receives.
    \item $hs_i$: feature determining if node $i$ sends message. 
    \item $hz_i$: feature determining if node $i$ outputs zero.
    \item $ho^+_i$ and $ho^-_i$: features determining the output of node $i$.
\end{itemize}
During the forwarding computations, $h_s$ and $h_z$ are constant. In practice, they can be the results of a zero weight combined with a constant bias. Their values in the two constructed models are shown in Table~\ref{table:h0}. The upcoming construction will ensure that, if $hs_i = k_s$, node $i$ does not send any message, and if $hz_i = k_z$, output of node $i$ will be zero. 

\begin{table}[ht]
\centering
\caption{The constant features of the TGNNs in $G_v$'s proof.}
\label{table:h0}
\scalebox{0.8}{
\begin{tabular}{|l|ccc|ccc|ccc|ccc|}\hline
\textbf{Node}            & \multicolumn{2}{c|}{\textbf{1}}      & \multicolumn{2}{c|}{\textbf{2}}      & \multicolumn{2}{c|}{\textbf{3}}      & \multicolumn{2}{c|}{\textbf{4}}      \\ \hline
\textbf{Hidden features} & \multicolumn{1}{c|}{$hs_1$} & \multicolumn{1}{c|}{$hz_1$} &  
                           \multicolumn{1}{c|}{$hs_2$} & \multicolumn{1}{c|}{$hz_2$} & 
                           \multicolumn{1}{c|}{$hs_3$} & \multicolumn{1}{c|}{$hz_3$} & 
                           \multicolumn{1}{c|}{$hs_4$} & \multicolumn{1}{c|}{$hz_4$}  \\ \hline
\textbf{TGNN $\Phi_1^v$}         & \multicolumn{1}{c|}{$k_s$}      & \multicolumn{1}{c|}{0}      & 
                           \multicolumn{1}{c|}{0}      & \multicolumn{1}{c|}{$k_z$}       & 
                           \multicolumn{1}{c|}{0}      & \multicolumn{1}{c|}{$k_z$}       & 
                           \multicolumn{1}{c|}{$k_s$}      & \multicolumn{1}{c|}{$k_z$}    \\ \hline
\textbf{TGNN $\Phi_2^v$}         & \multicolumn{1}{c|}{$k_s$}      & \multicolumn{1}{c|}{0}      & 
                           \multicolumn{1}{c|}{$k_s$}      & \multicolumn{1}{c|}{$k_z$}    & 
                           \multicolumn{1}{c|}{0}      & \multicolumn{1}{c|}{$k_z$}      & 
                           \multicolumn{1}{c|}{0}      & \multicolumn{1}{c|}{$k_z$}    \\ \hline
\end{tabular}
}
\end{table}

\begin{figure}[ht]
    \centering
    \begin{minipage}{0.39\linewidth}
        \centering
        \includegraphics[width=0.75\textwidth]{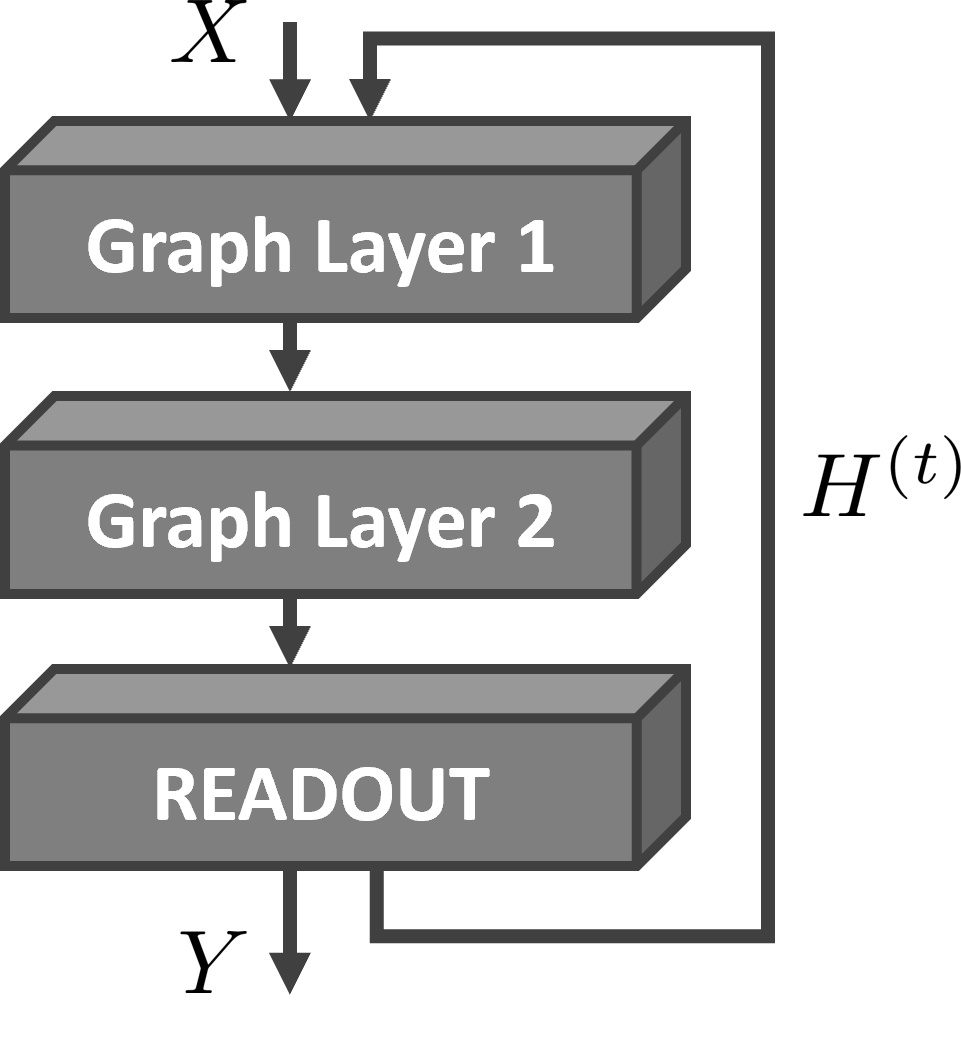} 
        \caption{The model.}
        \label{fig:model}
    \end{minipage}
    \begin{minipage}{0.6\linewidth}
        \centering
          \includegraphics[width=.85\linewidth]{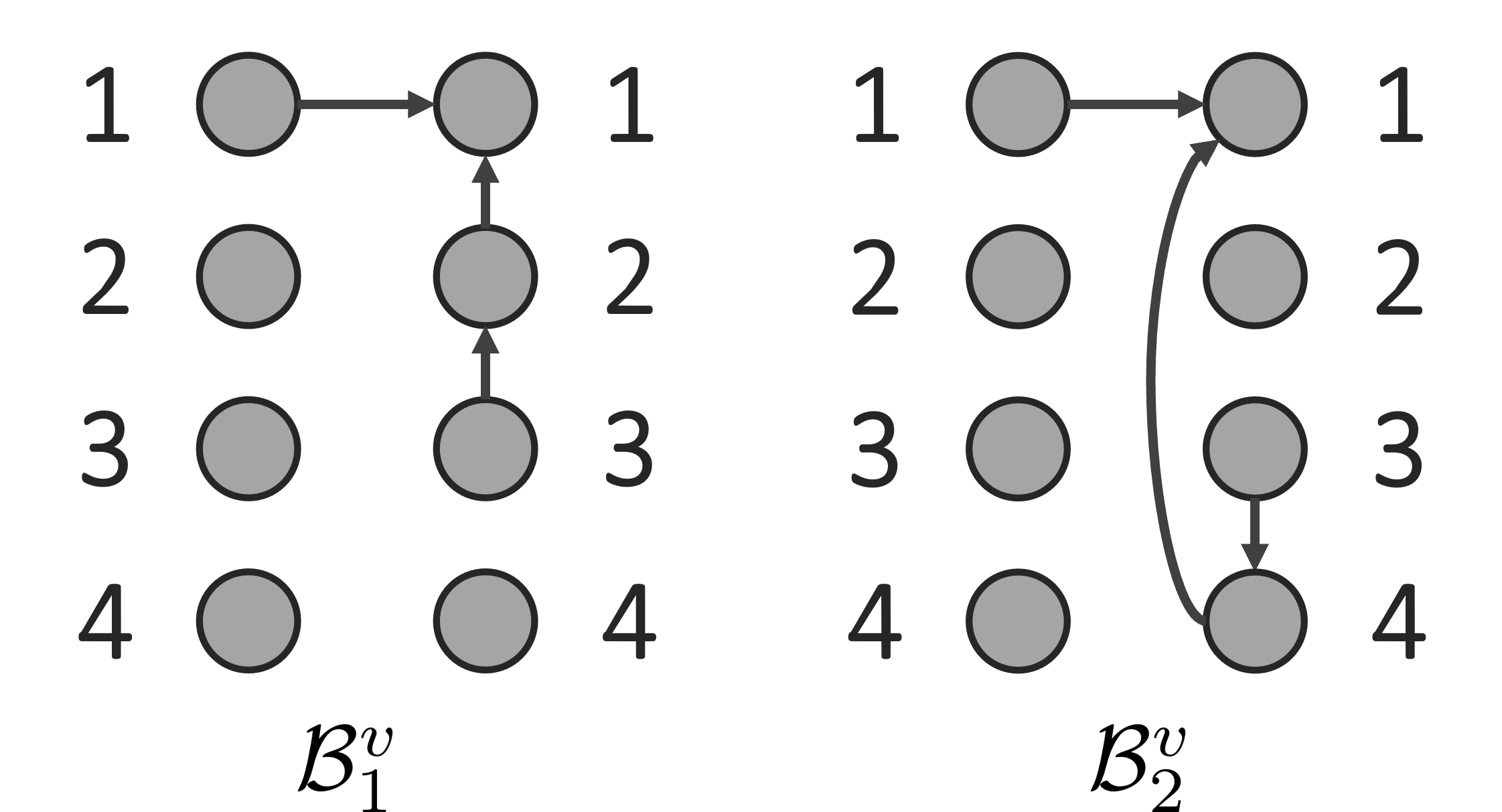}
        \caption{The DBNs explaining the two T-GNNs $\Phi_1^v$ and $\Phi_2^v$.}
        \label{fig:explanations}
    \end{minipage}\hfill
\end{figure}

Our proposed TGNN architecture has 2 graph layers followed by a READOUT layer (Fig.~\ref{fig:model}). By conventions, we use $l \in \{0,1,2\}$ to indicates the model's graph layers with $h_i^{(t,l=0)}$ refers to the model's input:
\begin{align}
    \bm h_i^{(t,l=0)} &= \left[X_i^{(t)}, H_i^{(t-1)}, *, *, 0, 0 \right]
    \label{eq:u0}
\end{align}
where $*$ means the features are pre-initialized by the model, i.e. $hs$ and $hz$. The temporal signal $H^{(t-1)}$ has the same dimension as the model's output $Y$, which is $\mathbb{R}^4$.

The two models have the trainable weights and biases such that the MSG and AGG functions are as follows:
    \begin{align}
        m_{ji}^{(t, l)} &= \textup{ReLU} \left(  hr_j^{(t, l-1)} - hs_j^{(t, l-1)} \right) \label{eq:m1}\\
        a_i^{(t, l)} &= \sum_{j \in \mathcal{N}_i} m_{ji}^{(t,l)} \label{eq:a1}
    \end{align}
for $l \in \{1,2\}$. This means, if $hs_j = k_s$ is large and kept unchanged, there is no message coming out of node $j$. Thus, the $a_i$ consists of messages only from node $j$ with $hs_j = 0$. 

Meanwhile, the UPD is a linear combination of previous hidden features and the aggregation $a_i$, followed by a ReLU: $
      \bm h_i^{(t,l)} = \textup{ReLU} \left( \bm w_h^\top {\bm h}_i^{(t,l-1)} + w_a a_i^{(t, l)}\right) $,
where the matrix $\bm w_h$ and $w_a$ are chosen such that:
\begin{align}
      &hr_i^{(t,l)} =   \textup{ReLU} \left(a_i^{(t, l)} \right), 
       ht_i^{(t,l)} =  \textup{ReLU} \left(ht_i^{(t,l-1)} \right) \label{eq:update}\\
        &hs_i^{(t,l)} =  \textup{ReLU} \left(hs_i^{(t,l-1)} \right), 
          hz_i^{(t,l)} =  \textup{ReLU} \left(hz_i^{(t,l-1)} \right) \nonumber\\
         &ho_i^{+(t,l)} =  \textup{ReLU} \left(a_i^{(t, l)} - ht_i^{(t, l-1)}  \right) \nonumber \\
         &ho_i^{-(t,l)} =  \textup{ReLU} \left(ht_i^{(t, l-1)} - a_i^{(t, l)} \right)
         \label{eq:uend}
\end{align}


With the above specification, it is clear that $h_t, h_s$ and $h_z$ are unchanged during the forwarding computation and $hr_i$ is the received signal at node $i$ (note that it is the input for $l=0$ based on Eq.~\ref{eq:u0}). By setting the READOUT as:
\begin{align}
    H_i^{(t)} &= \textup{ReLU} \left( ({hr_i + ht_i + ho_i^+ + ho_i^- })/{2} - hz_i\right) \label{eq:endreadout}
\end{align}
and assigning prediction of the model $Y$ to $ H^{(t=2)}$, we can show that the models' outputs satisfying Eq.~\ref{eq:task_node} (Appx.~\ref{appendix:node:forwarding}).

\medskip

\textbf{The Transparent Models of $\Phi_1^v$ and $\Phi_2^v$.}
We now examine the Transparent models $\mathcal{I}(\Phi_1^v(\bm X))$ and $\mathcal{I}(\Phi_2^v(\bm X))$. Fig.~\ref{fig:explanations} shows two DBNs whose variables represents the messages coming out of the model's nodes. Particularly, the variable of node $i$ at time $t$, denoted by $\mathcal{V}_i^t$, represents $m_{ij}^{(t,l)}$  ($\forall l, \forall j \in \mathcal{N}_i$) and the $H_i^{(t)}$. Our claim is the two DBNs can faithfully explain the two models when their inputs are bounded by the model's parameters, i.e. $K := \min \{ k_s, k_z\}$:

\begin{lemma} \label{lemma:dbn_explain}
The DBN $\mathcal{B}_1^v$ ($\mathcal{B}_2^v$) in Fig.~\ref{fig:explanations} can embed all information of the hidden features of TGNN $\Phi_1^v$ ($\Phi_2^v$ ) without any loss when the input signal is bounded by $K \vcentcolon = \min \{k_s, k_z\}$. Furthermore, the DBN is a minimal (Proof in Appx.~\ref{appendix:proof:dbn_explain}).
\end{lemma}

From Lemma~\ref{lemma:dbn_explain}, 
we have $  \mathcal{B}_1^v = \mathcal{I}(\Phi_1^v(\bm X))$ and $\mathcal{B}_2^v = \mathcal{I}(\Phi_2^v(\bm X)) $ for all $\bm X$ such that its entries are bounded by $K$. 

\medskip

\textbf{Unidentifiable Proof.} 
As the two DBNs contains different information, e.g. information about $\mathcal{V}_2$ and $\mathcal{V}_4$, we can state that the Transparent Models of $\Phi_1^v$ and $\Phi_2^v$ are different for some $\bm X$ bounded by $K$. Hence, explanation methods should be able to determine which DBN better describes each model, or at least returns distinctive explanations in explaining them. Unfortunately, in the next Lemma~\ref{lemma:inexplanability}, we show that for all such bounded $\bm X$, the outputs of the two TGNNs are exactly the same. The proof relies on the fact that both models satisfy Eq.~\ref{eq:task_node}.
 
\begin{lemma} \label{lemma:inexplanability}
For all $ \bm X $ such that $X_i^{(t)} \leq \min \{k_s, k_z\}$, we have $\Phi_1^v(\bm X) =  \Phi_2^v(\bm X) $ (Proof in Appx.~\ref{appendix:proof:inexplanability}).
\end{lemma}



By setting the domain $\mathcal{S}$ (Eq.~\ref{cond:same}) to be the set of inputs bounded by $K$,  Lemma~\ref{lemma:inexplanability} gives us that condition. We are now ready to show the unidentifiable result, which is stated in the below Theorem. Note that, we say a DBN $\mathcal{B}$ is the Transparent Model of $\Phi$ if $\mathcal{B}$ is a minimal DBN that can represent all communicating messages during the forward computation of $\Phi$.

\begin{theorem} \label{theorem:inexplanability_node}


For a TGNN $\Phi$ (Eq.~\ref{eq:tgnnlast}), denote $\mathcal{P} \vcentcolon= \{ ( X, A, \Phi(X, A) ) | X_i \leq K\}_{X} $, i.e. the set of Node-perturbation-response of $\Phi$ when the perturbations are bounded by $K$. Denote $g$ an arbitrary algorithm accepting $\mathcal{P}$ as inputs.  For any $K>0$ and for any $g$, there exists a $\Phi$ such that:
\begin{enumerate}
    \item For the interpretable domain of DBNs, the Transparent Model of $\Phi$ exists for all inputs in $\mathcal{P}$.
    \item {$g$ cannot determine the Transparent Model of $\Phi$.}
\end{enumerate}
\end{theorem}

\begin{proof}
We first set $k_s$ and $k_z$ (Table~\ref{table:h0}) to $K$. We then construct $\Phi_1^v$ and $\Phi_2^v$ as described from Eq.~\ref{eq:u0} to Eq.~\ref{eq:endreadout}. Denote $\mathcal{P}_1$ and $\mathcal{P}_2$ the sets of Node perturbation-response of $\Phi_1^v$ and $\Phi_2^v$, respectively. Note that, from Lemma~\ref{lemma:dbn_explain}, we have $\mathcal{B}_1^v$ and $\mathcal{B}_2^v$ are the Transparent Models of $\Phi_1^v$ and $\Phi_2^v$. 

Given a Node-perturbation-response, suppose $g$ returns either $\mathcal{B}_1^v$ or $\mathcal{B}_2^v$ (or equivalents claims on which DBN is more fit). As $\mathcal{P}_1$ is the same as $\mathcal{P}_2$ (Lemma~\ref{lemma:inexplanability}), the outputs of $g$ on the 2 perturbation-response sets must be the same. If, for example, $g(\mathcal{P}_1) = \mathcal{B}_1^v$, $g$ cannot {determine that $\mathcal{B}_2^v$ is the Transparent Model for}
$\Phi_2^v$ as $g(\mathcal{P}_2) = g(\mathcal{P}_1) = \mathcal{B}_1^v$. Thus, selecting $\Phi_2^v$ as $\Phi$ proves the Theorem.
\end{proof}


From the proof of Theorem~\ref{theorem:inexplanability_node}, we see that, even though $\mathcal{B}_1^v$ and $\mathcal{B}_2^v$ contain different information about how the messages propagate, all $g\in \mathcal{G}_v$ consider $\Phi_1^v$ and $\Phi_2^v$ the same. We then can conclude that Node-perturbation cannot identify which paths carrying out the model's predictions.

\section{Unidentifiable Proof for Edge-perturbation} \label{sect:edge}

Since the Unidentifiable Proof for Edge-perturbation shares a similar approach 
to that of Node-Perturbation and owing to space limit, this section highlights key ideas of the proof for Edge-perturbation class $\mathcal{G}_e$. The full proof is provided in Appx.~\ref{appendix:full_edge}. We show in our proof that removing edges from input graphs is not enough to identify all nodes contributing to a max operation conducted by the TGNNs. The intuition is, if the messages are gated by the features, edge perturbation does not reveal the sources of those messages. The unidentifiable results for GNNs are discussed in Sect.~\ref{sect:discussion}.

\textbf{The Training Task and the Models.} Our proof considers a graph of 3 nodes forming a line. The task (Fig.~\ref{fig:task_edge_top}) is to recognize the maximum positive inputs observed in node 2 and 3, and return result at node 1:
\begin{align}
 Y_1^{(t)} &= \max \left\{0, X_2^{(t')} \textup{ and } X_3^{(t')} , 0\leq t' \leq t \right\} \label{eq:task_edge}
\end{align}
The outputs on other nodes are zeros. 

We use the same architecture as in Sect.~\ref{sect:node} (Fig.~\ref{fig:model}) to construct two TGNNs named $\Phi_1^e$ and  $\Phi_2^e$. 
The hidden vector of each node has 5 main features, i.e. $\bm h_i  = [hr_i, ht_i, hs_i, hz_i, hl_i]$, and 6 additional features just to generate outputs. The only new feature in $\bm h_i$ is $hl_i$, which is constructed to be the lag version of $hr_i$. The models use the same MSG, AGG and UPD functions as described from Eq.~\ref{eq:m1} to Eq.~\ref{eq:uend}. The difference between the two models is only in node 3: while it sends message in $\Phi_1^e$ (set $hs_3 = 0$), it does not in $\Phi_2^e$ (set $hs_3 = k_s$). Meanwhile, the 6 additional features and the READOUT are constructed so that:
\begin{align*}
    H_1^{(t)} 
    & = \max \left \{  X_3^{(t)},  X_2^{(t)} , H_1^{(t-1)} \right\} 
\end{align*}
in $\Phi_1^e$. Given that, we can make $\Phi_1^e$ satisfies Eq.~\ref{eq:task_edge} by assigning the output $Y$ to $H^{(t)}$. On the other hand, as node 3 does not send messages in $\Phi_2^e$, it makes $ H_1^{(t)}$ the max of $X_2^{(t)}$ and $H_1^{(t-1)}$. This sets the output of $\Phi_2^e$ on node 1 to:
\begin{align}
 Y_1^{(t)} &= \max \left\{ 0 \textup{ and } X_2^{(t')} , 0\leq t' \leq t \right\} \label{eq:task_edge_2}
\end{align}
By comparing Eq.~\ref{eq:task_edge} to Eq.~\ref{eq:task_edge_2}, it is clear that $\Phi_1^e$ and $\Phi_2^e$ are different. However, when $X_2^{(t)} > X_3^{(t)}$, the responses of $\Phi_1^e$ and $\Phi_2^e$ are the same, which is stated in Lemma~\ref{lemma:inexplanability_edge}:
\begin{lemma} \label{lemma:inexplanability_edge}
For the task in Fig~\ref{fig:task_edge_top}, denote $\bar{A}$ the adjacency matrix obtained by either keeping the input adjacency matrix $A$ unchanged or by removing some edges. For any given $\bm X $ such that $X_i^{(t)} \leq \min \{k_s, k_z\}$ and $X_2^{(t)} > X_3^{(t)}$, we have $
    \Phi_1^e(\bm X, \bar{A}) =  \Phi_2^e(\bm X, \bar{A}) $ (Proof in Appx.~\ref{appendix:full_edge}). 
\end{lemma}

\textbf{The Transparent Models and Unidentifiable Proof.}
In the case of Node-perturbation, we associate the variable $\mathcal{V}_i^{t}$ in the DBN with the output messages and the propagating messages originated from node $i$. In Edge-perturbation, we can do the same by setting the messages $m_{ij}$ and $m_{ji}$ to 0 when the edge $(i,j)$ is removed. This conveniently saves us from the concerns regarding the actual perturbation schemes when using the interpretable domain as DBNs.

\begin{figure}[ht]
    \centering
    \begin{minipage}{0.47\linewidth}
        \centering
        \includegraphics[width=0.99\linewidth]{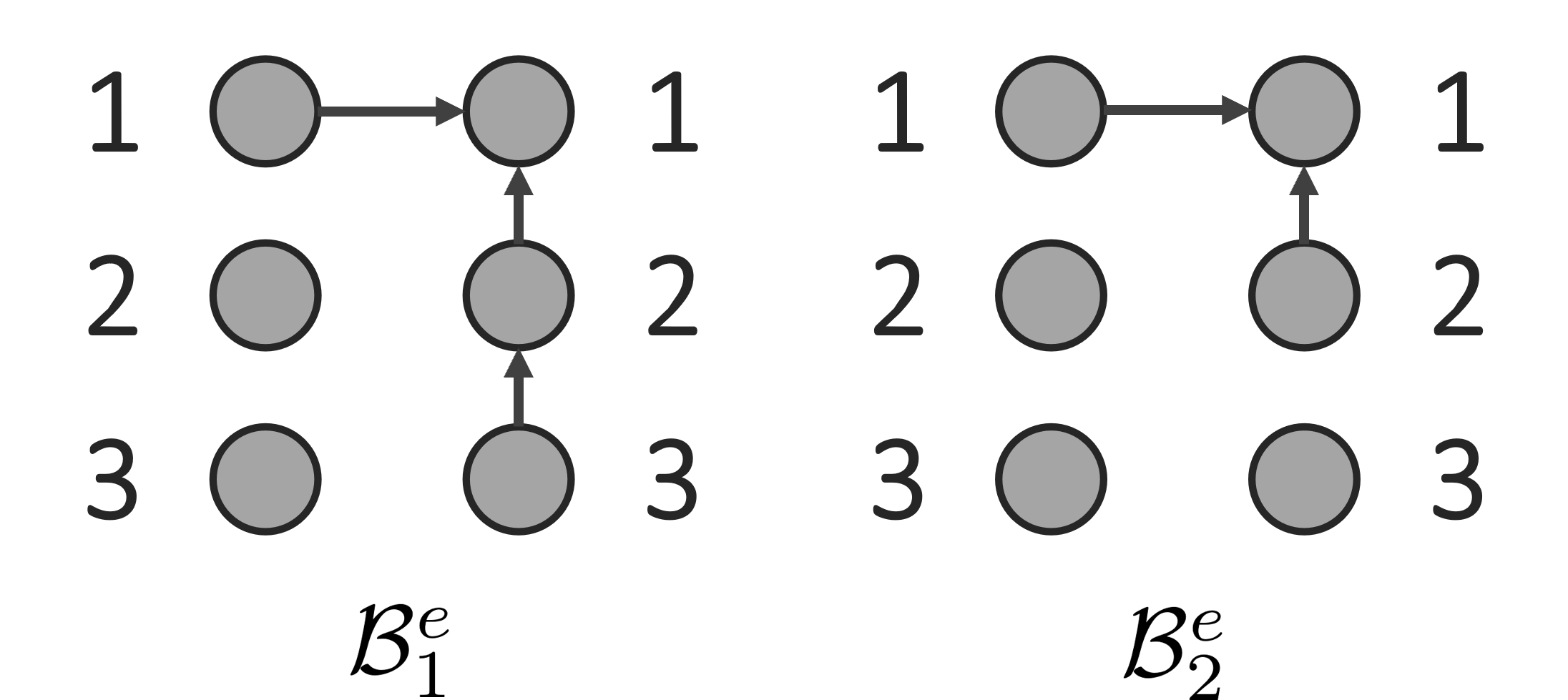} 
        \caption{The DBNs for Unidentifiable Proof of $\mathcal{G}_e$.}
        \label{fig:dbn_edge}
    \end{minipage} 
    \hfill
    \begin{minipage}{0.47\linewidth}
        \centering
          \includegraphics[width=0.99\linewidth]{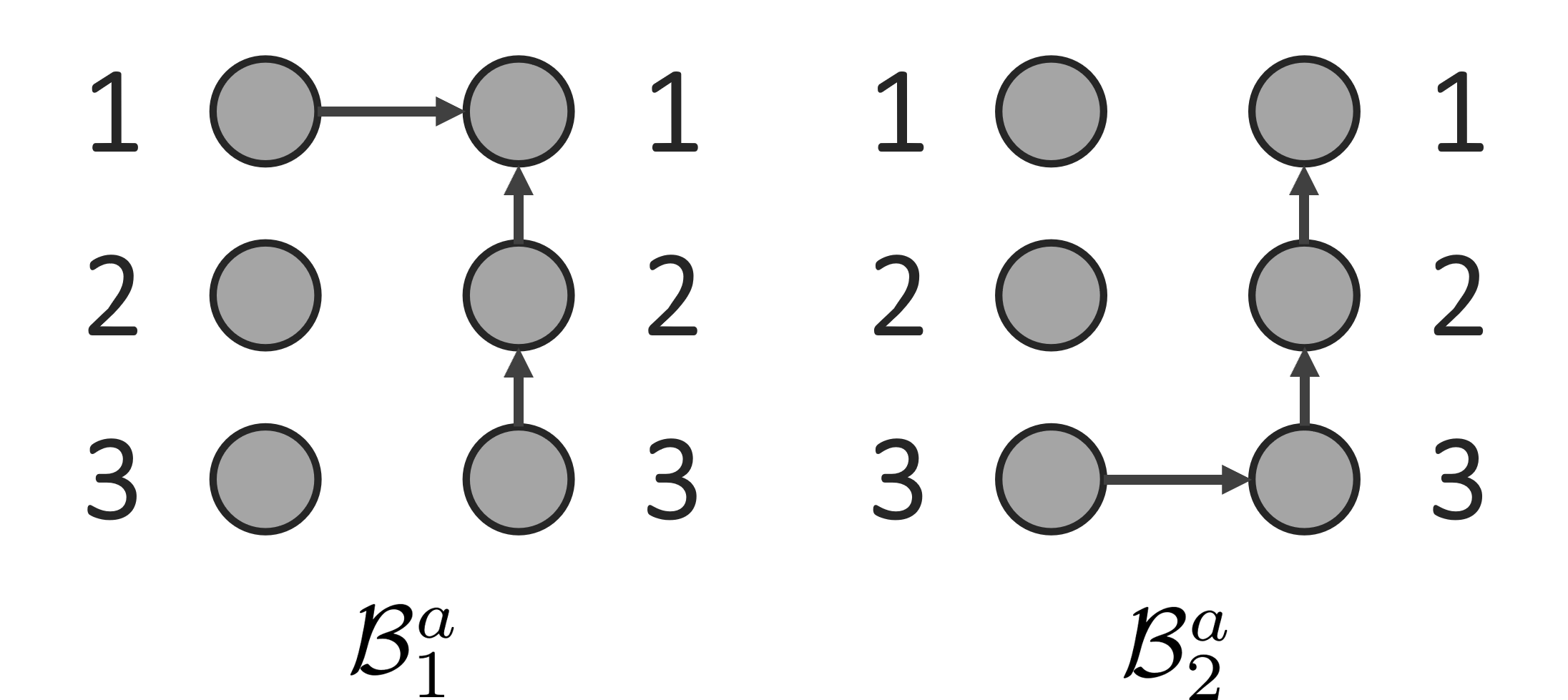}
        \caption{The DBNs for Unidentifiable Proof of $\mathcal{G}_a$.}
        \label{fig:dbn_node_edge}
    \end{minipage}\hfill
\end{figure}

As $\Phi_1^e$ is different with $\Phi_1^v$ only in node 4 and the additional content in the propagating messages, it follows that $\mathcal{B}_1^e$ (Fig.~\ref{fig:dbn_edge}) is the Transparent Model of $\Phi_1^e$ for $\bm X$ bounded by $\min \{k_s, k_z\}$. We write $\mathcal{B}_1^e =\mathcal{I}(\Phi_1^e)$. Note that even when $X_2^{(t)} > X_3^{(t)}$, $m_{21}^{(t,l=1)}$ is determined by $X_3^{(t)}$. Thus, the edge between $\mathcal{V}_{3}^t$ and $\mathcal{V}_{2}^t$ in $\mathcal{B}_1^e$ is necessary. Regarding
$\Phi_2^e$, as it is just $\Phi_1^e$ with node 3 disconnected,  $\mathcal{B}_2^e$ (Fig.~\ref{fig:dbn_edge}) is the Transparent Model of $\Phi_2^e$, i.e. $\mathcal{B}_2^e =\mathcal{I}(\Phi_2^e)$.  The above arguments combined with Lemma~\ref{lemma:inexplanability_edge} give us the following Theorem about the Unidentifiable Proof for Edge-perturbation:

\begin{theorem} \label{theorem:inexplanability_edge}

For a TGNN $\Phi$ (Eq.~\ref{eq:tgnnlast}), denote $\mathcal{P} \vcentcolon= \{ ( X, \bar{A}, \Phi(X, \bar{A})) | X_i \leq K\}_{\bar{A}} $, i.e. the set of Edge-perturbation-response of $\Phi$ where $\bm X$ are fixed and bounded by $K$, and $\bar{A}$ is defined as in Lemma~\ref{lemma:inexplanability_edge}. Denote $g$ an arbitrary algorithm accepting $\mathcal{P}$ as inputs.  For any $K>0$ and any $g$, there exists a $\Phi$ satisfying the two conditions in Theorem~\ref{theorem:inexplanability_node} (Proof in Appx.~\ref{appendix:full_edge}).

\end{theorem}

The proof of Theorem~\ref{theorem:inexplanability_edge} is obtained by replacing $\Phi_1^e$, $\Phi_2^e$, $\mathcal{B}_1^e$ and $\mathcal{B}_2^e$ with $\Phi_1^v$, $\Phi_2^v$, $\mathcal{B}_1^v$ and $\mathcal{B}_2^v$ in that of Theorem~\ref{theorem:inexplanability_node}. Since $\mathcal{B}_1^e$ and $\mathcal{B}_2^e$ have different set of variables connected to node 1, we can conclude Edge-perturbation cannot reliably identify all components contributing to the prediction.

\section{Unidentifiable Proof for Node-and-Edge Perturbation in TGNN} \label{sect:node_edge}

This section provides the Unidentifiable Proof for the Node-and-Edge-perturbation class $\mathcal{G}_a$ in TGNNs. The proof demonstrates that perturbing both nodes and edges does not help identify which nodes carry out the temporal aggregation in the TGNNs. 

\textbf{The Training Task.}
The TGNNs operate on the same line-graph as in Sect.~\ref{sect:edge}. The task (Fig.~\ref{fig:task_node_edge_top}) is to recognize the maximum positive inputs observed in node 3 and return the result at node 1:
\begin{align}
 Y_1^{(t)} &= \max \{0 \textup{ and } X_3^{(t')} , 0\leq t' \leq t \}  \label{eq:spec}
\end{align}
The outputs on other nodes are zeros. This proof constructs 2 TGNNs whose internal behaviors are captured by the DBNs shown in Fig.~\ref{fig:dbn_node_edge}.  The main difference of this proof compared to previous proofs is that the models involve temporal messages of nodes other than the prediction node 1.


\textbf{The Models.} We use the same architecture as in Fig.~\ref{fig:model}. The hidden feature vectors have 7 features, i.e. $\bm h_i  = [hr_i, ht_i, hs_i, hz_i, ho, ho^+_i, ho^-_i]$. Except from the newly introduced $ho_i$, all features have the same meaning as described in Sect.~\ref{sect:node}. The two constructed models, called $\Phi_1^a$ and $\Phi_2^a$, have different MSG functions, READOUT functions and hidden constant features, i.e. $hs_i$ and $ho_i$. The constant features for the two models are shown in Table~\ref{table:h2}.


The MSG, AGG, UPD and READOUT of $\Phi_1^a$ are as specified from Eq.~\ref{eq:m1} to \ref{eq:endreadout}. Since $ho_i$ in $\Phi_1^a$ is just a dummy variable, the output of $\Phi_1^a$ satisfies Eq.~\ref{eq:spec} since $\Phi_1^a$ is $\Phi_1^v$ without node 4 and Eq.~\ref{eq:spec} is the same as Eq.~\ref{eq:task_node}.

\begin{table}[ht]
\centering
\caption{The constant features of the TGNNs in $G_a$'s proof.}
\label{table:h2}
\scalebox{0.8}{
\begin{tabular}{|l|ccc|ccc|ccc|}\hline
\textbf{Node}            & \multicolumn{3}{c|}{\textbf{1}}      & \multicolumn{3}{c|}{\textbf{2}}      & \multicolumn{3}{c|}{\textbf{3}}          \\ \hline
\textbf{Features} & \multicolumn{1}{c|}{$hs_1$} & \multicolumn{1}{c|}{$hz_1$} &   \multicolumn{1}{c|}{$ho_1$} &
                           \multicolumn{1}{c|}{$hs_2$} & \multicolumn{1}{c|}{$hz_2$} & \multicolumn{1}{c|}{$ho_2$} &
                           \multicolumn{1}{c|}{$hs_3$} & \multicolumn{1}{c|}{$hz_3$} & \multicolumn{1}{c|}{$ho_3$}   \\ \hline
\textbf{TGNN $\Phi_1^a$}         & \multicolumn{1}{c|}{$k_s$}      & \multicolumn{1}{c|}{0}      & \multicolumn{1}{c|}{0}      & 
                           \multicolumn{1}{c|}{0}      & \multicolumn{1}{c|}{$k_z$}       & \multicolumn{1}{c|}{0}      & 
                           \multicolumn{1}{c|}{0}      & \multicolumn{1}{c|}{$k_z$}   & \multicolumn{1}{c|}{0} \\ \hline
\textbf{TGNN $\Phi_2^a$}         & \multicolumn{1}{c|}{$k_s$}      & \multicolumn{1}{c|}{0}      & \multicolumn{1}{c|}{0}      & 
                           \multicolumn{1}{c|}{0}      & \multicolumn{1}{c|}{$k_z$}    & \multicolumn{1}{c|}{$k_z$}      & 
                           \multicolumn{1}{c|}{$k_s$}      & \multicolumn{1}{c|}{$k_z$} & \multicolumn{1}{c|}{$k_z$}   \\ \hline
\end{tabular}
}
\end{table}

In $\Phi_2^a$, we use $hz_i$ to control the temporal messages $H^{(t)}$ and the newly introduced $ho_i$ is to control the output $Y$.
The model uses the same AGG and UPD functions as specified from Eq.~\ref{eq:a1} to \ref{eq:uend}. The update rule of $ho_i$ is $ho_i^{(t,l)} =  \textup{ReLU} (ho_i^{(t,l-1)} ) $.
The MSG function; however, has the trainable weight $ \bm w_m$ such that:
    \begin{align}
        m_{ij}^{(t, l)} &= \textup{ReLU} \left(  \bm w_m^\top \bm h_i^{(t,l-1)} \right) \label{eq:mdev}
    \end{align}
where $
    \bm w_m^\top \bm h_i  = (hr_i + ht_i+ho_i^+ + ho_i^-)/2 - hs_i$. Here, all variables except $\bm w_m$ have temporal index $t$ and layer index $l$. The final difference in $\Phi_2$ compared to $\Phi_1$ is its READOUT as we set $H_i^{(t)} = \textup{ReLU} \left( hr_i - hz_i\right)$ and $Y_i = \textup{ReLU} \left( hr_i - ho_i\right)$. Appx.~\ref{appendix:nodeedge:forwarding} shows that $\Phi_2^a$ fulfills Eq.~\ref{eq:spec}.

\medskip

\textbf{The Transparent Models of $\Phi_1^a$ and $\Phi_2^a$.} As  $\Phi_1^a$ is $\Phi_1^v$ without node 4, we have the $\mathcal{B}_1^a$, which is $\mathcal{B}_1^v$ without variables for node 4, is the Transparent Model of $\Phi_1^a$. We write $\mathcal{B}_1^a = \mathcal{I}(\Phi_1^a(\bm X))$ for $\bm X$ bounded by $\min \{k_s, k_z \}$. 

On the other hand, the transparent model of $\Phi_2^a$ can be shown to be the DBN $\mathcal{B}_2^a$ in Fig.~\ref{fig:dbn_node_edge} by the following Lemma:

\begin{lemma} \label{lemma:dbn_explain_2}
The DBN $\mathcal{B}_2^a$ in Fig.~\ref{fig:dbn_node_edge} can embed all information of the hidden features of TGNN $\Phi_2^a$  without any loss when the input signal is bounded by $K \vcentcolon = \min \{k_s, k_z\}$. Furthermore, the DBN is a minimal (Proof in Appx.~\ref{appendix:proof:dbn_explain_2}).
\end{lemma}



Since the two DBNs in Fig.~\ref{fig:dbn_node_edge} contain different information, e.g. different set of independent variables,  Lemma~\ref{lemma:dbn_explain_2} allows us to claim $
     \mathcal{I}(\Phi_1^a(\bm X)) \neq  \mathcal{I}(\Phi_2^a(\bm X)) $
for some $\bm X$ bounded by $K$.

\textbf{Unidentifiable Proof.} Similar to previous proofs, we show that, for all $\bm X$ bounded by $K$ and a valid adjacency matrix, the outputs of the two constructed models are the same, which is stated in the following Lemma:
 
\begin{lemma} \label{lemma:inexplanability_nodeedge}
For the training task in Fig~\ref{fig:task_node_edge_top}, denote $\bar{A}$ the adjacency matrix obtained by either keeping the input adjacency matrix $A$ unchanged or by removing some edges from $A$.
For all $ \bm X $ such that $X_i^{(t)} \leq \min \{k_s, k_z\}$, we have $
    \Phi_1^a(\bm X, \bar{A}) =  \Phi_2^a(\bm X, \bar{A})$ (Proof in Appx.~\ref{appendix:inexplanability_nodeedge}). 
\end{lemma}



We are now ready to state the Unidentifiable Proof for the class of Node-and-Edge-perturbation:

\begin{theorem} \label{theorem:inexplanability_node_edge}
For a TGNN $\Phi$ (Eq.~\ref{eq:tgnnlast}), denote $\mathcal{P} \vcentcolon= \{ ( X, \bar{A}, \Phi(X, \bar{A}) ) | X_i \leq K\}_{X \in \mathcal{X}, \bar{A}}$, i.e. the set of Node-and-Edge-perturbation-response of $\Phi$ where $\bm X$ are fixed and bounded by $K$, and $\bar{A}$ is defined as in Lemma~\ref{lemma:inexplanability_nodeedge}. Denote $g$ an arbitrary algorithm accepting $\mathcal{P}$ as inputs.  For any $K>0$ and any $g$, there exists a $\Phi$ satisfying the two conditions in Theorem~\ref{theorem:inexplanability_node} (Proof in Appx.~\ref{appendix:theorem_nodeedge}).
\end{theorem}

As in previous proofs, the proof of Theorem~\ref{theorem:inexplanability_node_edge} also shows Node-and-Edge-perturbation cannot differentiate $\Phi_1^a$ with $\Phi_2^a$, whose Transparent Models are two DBNs with different temporal information. Thus, we can conclude Node-and-Edge-perturbation cannot identify the model's components conducting the temporal aggregation.

\section{Experiments} \label{sect:exp}

\begin{figure*}[ht]
		\centering
         \begin{subfigure}{0.49\linewidth}
          \centering
          \includegraphics[height=3.9cm]{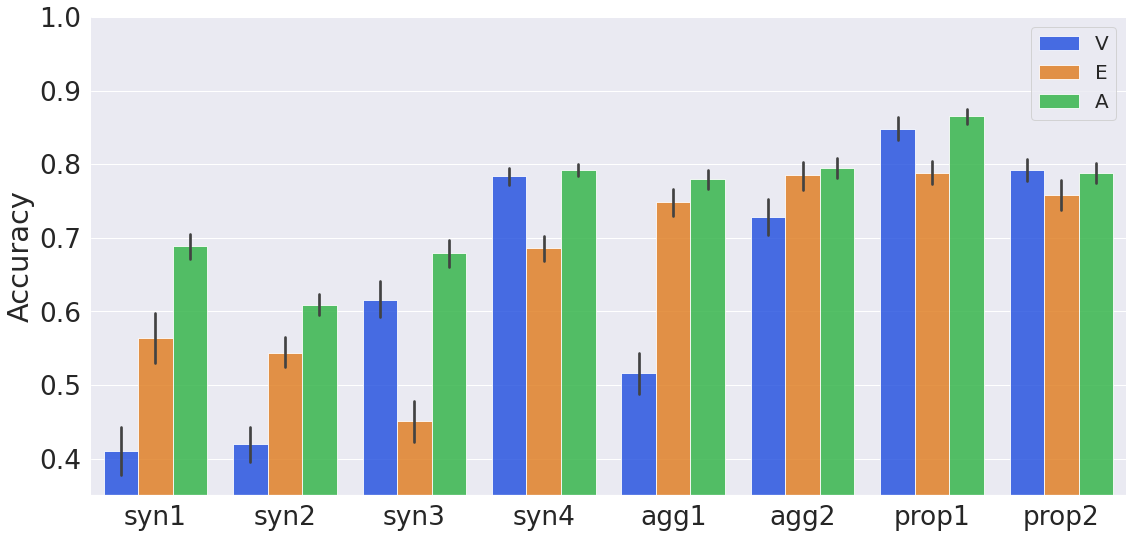}
          \caption{GCN}
                \label{fig:gcn_gnnex}
        \end{subfigure}
      \begin{subfigure}{0.49\linewidth}
          \centering
          \includegraphics[height=3.9cm]{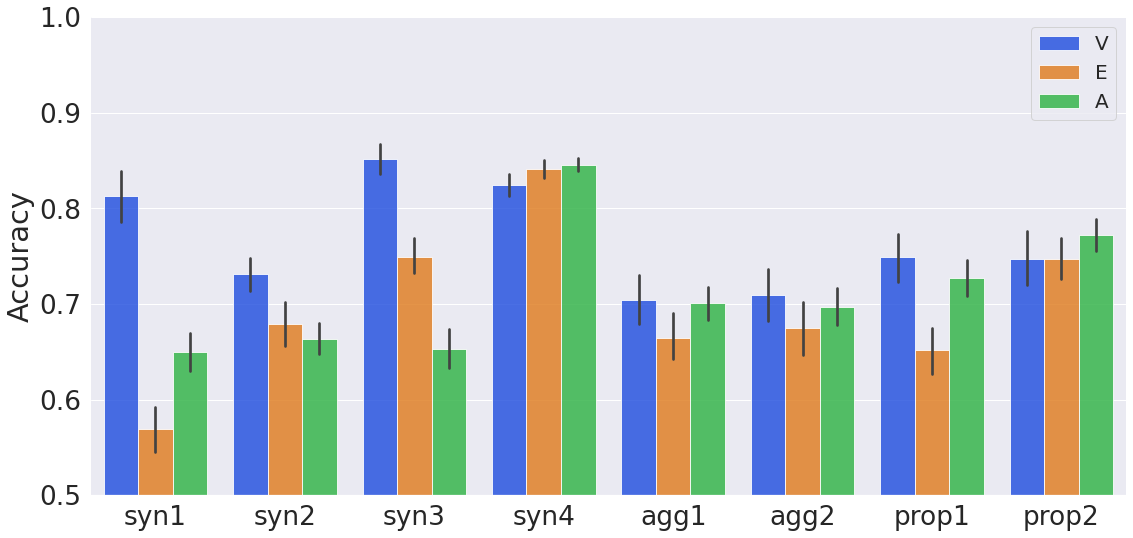}
          \caption{GIN}
                \label{fig:gin_gnnex}
        \end{subfigure}
        \caption{Accuracy of perturbation schemes on synthetic datasets. V, E and A stands for Node-perturbation, Edge-perturbation and Node-and-Edge-perturbation. The error bar shows the 95\% confidence interval of the results.}
        \label{fig:exp_syn}
\end{figure*}

\begin{figure}[ht]
		\centering
         \begin{subfigure}{.48\linewidth}
          \centering
          \includegraphics[height=3.5cm]{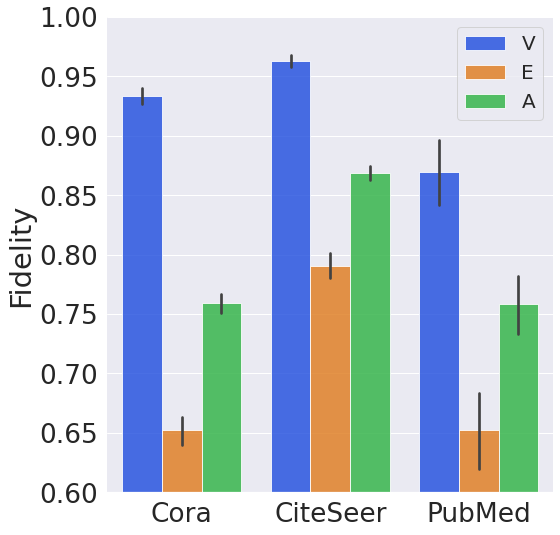}
          \caption{GCN}
                \label{fig:gcn_gnnexr}
        \end{subfigure}
        \begin{subfigure}{.48\linewidth}
          \centering
          \includegraphics[height=3.5cm]{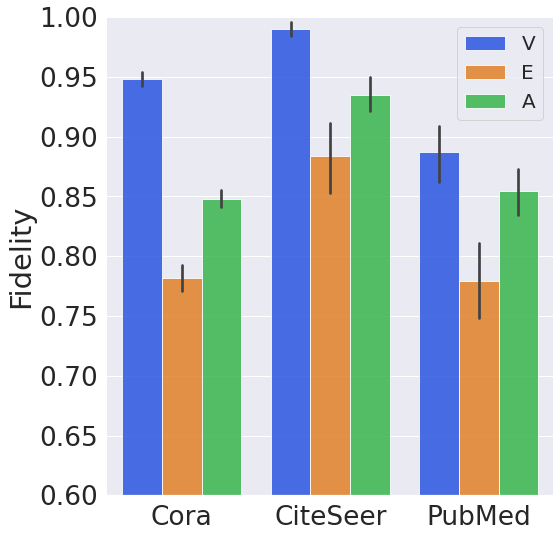}
          \caption{GAT}
                \label{fig:gat_gnnexr}
        \end{subfigure}
        \caption{Fidelity of perturbation schemes on real-world datasets. The notations have the same meaning as in Fig.~\ref{fig:exp_syn}.}
        \label{fig:exp_real}
\end{figure}

\begin{figure*}[ht]
		\centering
         \begin{subfigure}{.245\linewidth}
          \centering
          \includegraphics[width=\linewidth]{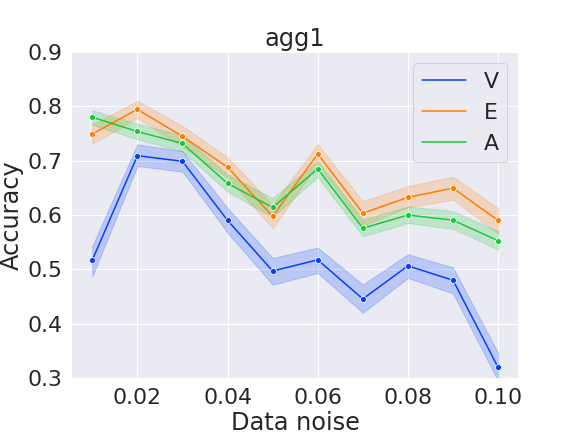}
        \end{subfigure}
         \begin{subfigure}{.245\linewidth}
          \centering
          \includegraphics[width=\linewidth]{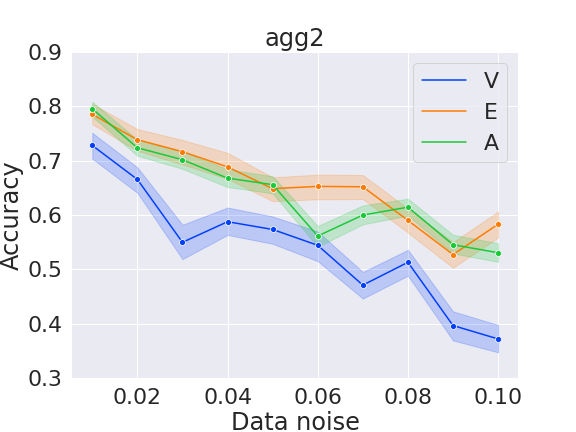}
        \end{subfigure}
      \begin{subfigure}{.245\linewidth}
          \centering
          \includegraphics[width=\linewidth]{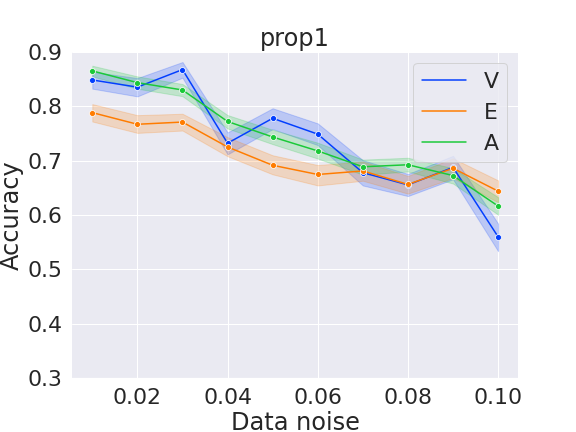}
        \end{subfigure}
         \begin{subfigure}{.245\linewidth}
          \centering
          \includegraphics[width=\linewidth]{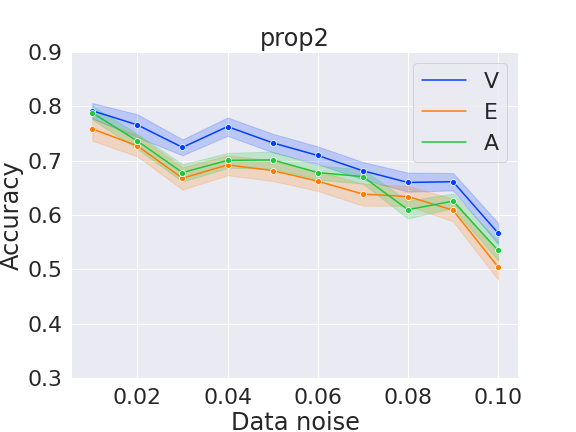}
        \end{subfigure}
        \caption{Accuracy in explaining GCN on noisy synthetic datasets. The notations have the same meaning as in Fig.~\ref{fig:exp_syn}.}
        \label{fig:exp_syn_gcn_noisy}
\end{figure*}

\begin{figure*}[ht]
		\centering
         \begin{subfigure}{.245\linewidth}
          \centering
          \includegraphics[width=\linewidth]{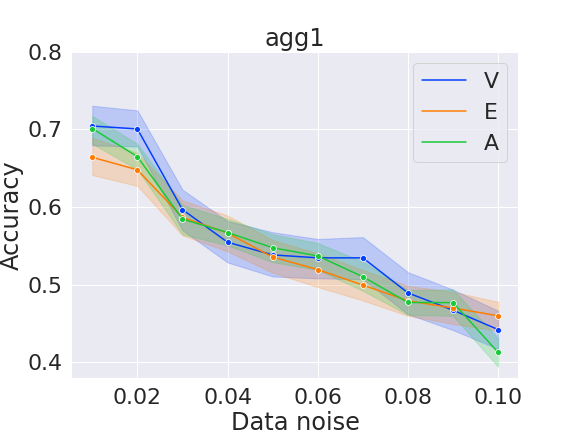}
        \end{subfigure}
         \begin{subfigure}{.245\linewidth}
          \centering
          \includegraphics[width=\linewidth]{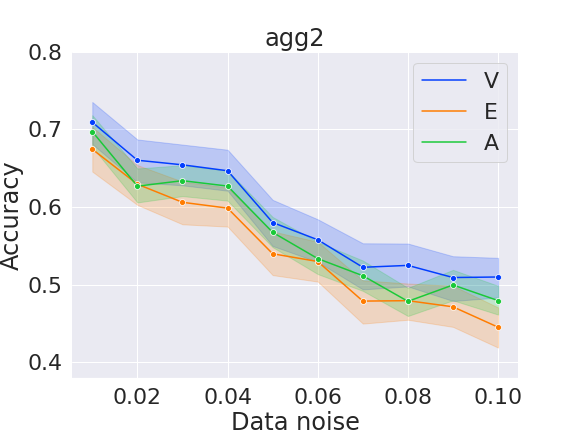}
        \end{subfigure}
      \begin{subfigure}{.245\linewidth}
          \centering
          \includegraphics[width=\linewidth]{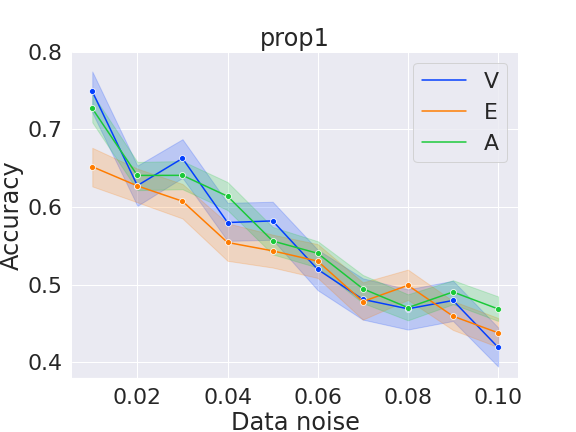}
        \end{subfigure}
         \begin{subfigure}{.245\linewidth}
          \centering
          \includegraphics[width=\linewidth]{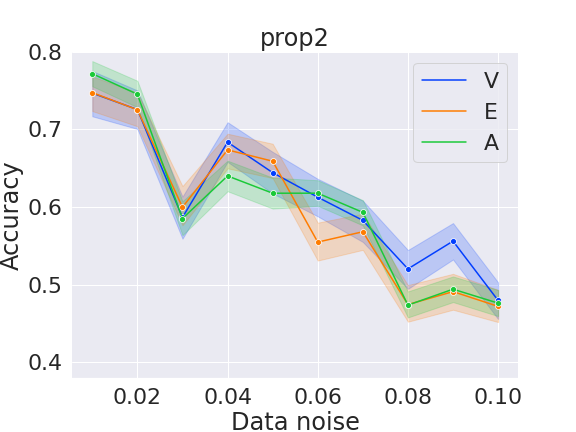}
        \end{subfigure}
        \caption{Accuracy in explaining GIN on noisy synthetic datasets. The notations have the same meaning as in Fig.~\ref{fig:exp_syn}.}
        \label{fig:exp_syn_gin_noisy}
\end{figure*}

This section provides our experiments examining the impact of perturbation schemes on the performance of explanation methods in synthetic and real-world datasets. 

{\bf Models and Explainers.} Since there is no existing explainers designed for general TGNNs, the models we study in our experiments are GCN, GIN and GAT in the node classification tasks. We train all models with Adam optimizer~\cite{adam_opt} with learning rate of 0.001. For synthetic experiments, we use the GCNs and GINs provided by \cite{funke2021zorro} and achieve at least 85\% test-set accuracy. We need to modify the GNNs to 4 layers so that they can carry out some new synthetic tasks. For the real-world datasets, we use the 2-layers GCN and GAT with accuracy at least 70\%, which is consistent with results reported in the original papers~\cite{Kipf2016iclr}. We do not conduct synthetic experiments for GAT and real-world experiments for GIN because we cannot train those models to competitive accuracy. We choose to study the GNNExplainer~\cite{Ying_NIPS2019} because that is the only black-box method supporting all perturbation schemes, which is essential for fair comparison. 


{\bf Synthetic experiments.} We follow the setting of~\cite{Ying_NIPS2019}, in which the synthetic datasets are constructed. Each input graph is a combination of a base graph and a set of motifs, whose details are shown in Table~\ref{synparameter}. We reuse four datasets of that previous work (called syn 1 to syn 4) and construct 4 new synthetic datasets. The purpose of the 4 new datasets is to analyze the impact of training tasks on the perturbation schemes. In the \textit{Aggregation} datasets, i.e. Agg 1 and Agg 2, the task requires all nodes to aggregate their information. In fact, each node in a motif has a role, which is one-hot encoded in the node's features. Each node also has another set of random features. The labels of all nodes in the motif is the label of the node with maximum sum of random features. On the other hand, in the \textit{Propagation} datasets, Prop 1 and Prop 2, information needs to propagate between some pairs of nodes. Particularly, the node's label is determined by the role randomly encoded in another fix position in the motif. In order for the models to correctly predict the labels, the features encoded on those fix positions need to propagate in the model to the target nodes. 

\begin{table}[h]
\caption{Parameters of synthetic datasets.}\label{synparameter}
\scalebox{0.79}{
\begin{tabular}{|c|l|l|l|}
\hline
\multicolumn{1}{|l|}{\textbf{Dataset}} & \textbf{Base}              & \textbf{Motifs}               & \textbf{Features} \\ \hline
Syn 1                                  & 300-node BA graph          & 80 5-node house-shaped   & Constant                 \\ \hline
Syn 2                                  & 300-node BA graph          & 80 9-node grid-shaped    & Constant                 \\ \hline
Syn 3                                  & Tree with height 8 & 60 6-node cycle-shaped   & Constant                 \\ \hline
Syn 4                                  & Tree with height 8 & 80 9-node grid-shaped    & Constant                 \\ \hline
Agg 1                                  & 300-node BA graph          & 80 6-node cycle-shaped  & From label                \\ \hline
Agg 2                                 & 300-node BA graph          & 80 6-node tree-shaped  & From label                \\ \hline
Prop 1                                  & 300-node BA graph          & 80 6-node cycle-shaped  & From label                \\ \hline
Prop 2                                 & 300-node BA graph          & 80 6-node tree-shaped  & From label                 \\ \hline
\end{tabular}
}
\end{table}

 Since the ground-truth explanations are given in synthetic dataset, we use accuracy as an evaluation metric for perturbation methods. Note that for Node-perturbation and Edge-perturbation, the explanations are in form of nodes and edges respectively. The accuracy is computed based on whether the selected nodes or edges are in the ground-truth motif. The accuracy of Node-and-Edge-perturbation are the average of the two accuracy. Fig.~\ref{fig:exp_syn} shows the accuracy of GNNExplainer with 3 different perturbation schemes on synthetic datasets.
 
 The first observation is that perturbation schemes have significant impact on the accuracy of the explainer. For GCN, we can see that node-perturbation is worse than edge-perturbation in syn1, syn2 and the Aggregation tasks. On the other hand, perturbing all features dominates other methods in all experiments. For GIN, edge-perturbation shows to be the worst in 6 out of 8 experiments. Perturbing node 
 shows to be the best method in most experiments.

{\bf Real-world experiments.} For real-world experiments, we use the Cora, Citeseer and Pubmed citation datasets~\cite{sen2018}. With the absence of ground-truth explanations, we use the \textit{fidelity} metric~\cite{funke2021zorro} to evaluate explanations. Intuitively, the metric is the probability that the prediction changes when the nodes included in the an explanation are fixed. In our experiments, the number of nodes in the explanation is 5. Since the fidelity metric is defined for nodes only, we convert the edge score of edge-perturbation to node score and compute the fidelity, i.e. the score of a node is the score of the maximum edge connecting to it.
 Experiments on real-world datasets shown in Fig.~\ref{fig:exp_real} also demonstrate clear impact of perturbation methods on the performance of explainer. It is interesting to see that perturbing node gives the highest fidelity in both models. 
 This implies perturbing all might not always be optimal.

\textbf{Stress Tests.} To test the robustness of results obtained in synthetic experiments (Fig.~\ref{fig:exp_syn}), we perturb the synthetic data, i.e. changing the probability of inserting and removing edges from the original graph, and conduct similar evaluations on those synthetic datasets. Fig.~\ref{fig:exp_syn_gcn_noisy} and Fig.~\ref{fig:exp_syn_gin_noisy} show the accuracy of perturbations methods in 4 new synthetic datasets with different noise levels. Each point shown in the plots correspond to one dataset and one model trained on that dataset. The fluctuation of the results came from the randomness of the data and the trained models. Nevertheless, we can observe that, given a task and a model architecture, the relative ordering of performance of perturbation schemes hold in the experiments. 
This implies the selection of optimal perturbation schemes should consider the model's architecture and the task at hand, which is currently ignored by many explanation methods.


\section{Discussion and Conclusion} \label{sect:discussion}

This work studies the fundamental limit of different perturbation explanation methods in explaining black-box TGNNs. We have shown that there are key information on how the TGNNs generate their predictions that cannot be identified by some given classes of explanation methods. We now further point out several interesting implications of our theoretical results.

\textbf{Theorem~\ref{theorem:inexplanability_node} and \ref{theorem:inexplanability_edge} for GNNs.} The  Unidentifiable Proofs for Node-perturbation and Edge-perturbation explanation methods can be applied directly to GNNs by dropping the feedback loop of $H^{(t)}$ (Fig.~\ref{fig:model}). This modification will just set $H_i^{(t-1)}$ in Eq.~\ref{eq:u0} to zeros. Note that in GNNs, we need to drop the temporal dimension in the training task and the interpretable domain, i.e use BNs instead of DBNs. Appx.~\ref{appendix:UP_for_gnns} provides more details on those proofs.

\textbf{Theorem~\ref{theorem:inexplanability_node_edge} for GNNs?} Our proof cannot readily apply to the case of GNNs because the two constructions will have the same Transparent Model, i.e. $\mathcal{I}(\Phi_1^a) = \mathcal{I}(\Phi_2^a)$. 

\textbf{What practical models are applicable by our Unidentifiable Proofs?} As our analysis use the most basic constructions of the TGNNs, in which the graph layers are as elementary as possible, our Unidentifiable Proofs in Sects.~\ref{sect:node}, \ref{sect:edge} and \ref{sect:node_edge} are applicable to all versions of the TGNNs found in~\cite{zhao2019t,min2021stgsn,mohamed2020social}. 

Our results in Sects.~\ref{sect:node} and \ref{sect:edge} are also applicable to many modern variants of GNNs including ChebNets~\cite{Defferrard2016}, GCN~\cite{Kipf2016iclr}, GraphSage \cite{GraphSage}, GAT~\cite{velickovic2018graph} and GIN~\cite{xu2018how}.

\textbf{Usage of other interpretable domains and Transparent Models (not DBNs).} Unidentifiable results can be obtained by others interpretable domains as long as (i) the Transparent Models of the constructed models can be identified (similar to Lemma~\ref{lemma:dbn_explain} and \ref{lemma:dbn_explain_2})  and (ii) they contain meaningful information that help establish the unidentifiable information. While condition (i) requires the domain to have strong expressive power, condition (ii) requires the domain's members to be somewhat interpretable. We find DBN is a balance choice for the analysis of TGNN.

\textbf{What do Theorem~\ref{theorem:inexplanability_node} and \ref{theorem:inexplanability_edge} imply about the reliability of existing explanation methods for GNNs?} Existing explanation methods have been successfully identify many important features contributing to the predictions; however, the results are 
still limited. Our results establish a fundamental limit  of perturbation-based explanation methods.

For example, Theorem~\ref{theorem:inexplanability_node} implies explanations obtained by only perturbing nodes cannot reliably inform us the paths determining the predictions. For the case of the two constructed $\Phi_1^v$ and $\Phi_2^v$, the contributions of node 2 and node 4 will always be considered equal by all Node-perturbation methods. This means both will be included or discarded by the explainers, even when the actual messages are only transmitted through one of them. Thus, Node-perturbation methods bound to commit false positive or false negative.



\textbf{What does Theorem~\ref{theorem:inexplanability_node_edge}  imply about the design of explanation methods for TGNNs?} Even though the Theorem states that the Node-and-Edge perturbation methods cannot identify the temporal component of the model, it does not mean there is nothing we can do to tackle this challenging problem. Careful readers might realize that one key aspect of our proof is based on the fact that removing an edge in the input graph will disconnect that connection at all rounds of temporal computations. If there is a mechanism to remove edge only at some temporal computations, it is possible to differentiate $\Phi_1^a$ from  $\Phi_2^a$, which is crucial to identify whether node 1 or node 3 conducts the temporal aggregation. In other words, \textit{temporal perturbation} might be something we need to explain TGNNs more faithfully. 
\bibliography{minhbibfile}




\clearpage

\appendix

\section*{Technical Appendix}

\textbf{Appendix Outline} 

\begin{itemize}
    \item Appendix~\ref{appendix:prelim_bn}: preliminaries of Bayesian Networks and Dynamic Bayesian Networks.
    \item Appendix~\ref{appendix:node:forwarding}: the analysis of the forwarding computations of models constructed in Sect.~\ref{sect:node}.
    \item Appendix~\ref{appendix:proof:dbn_explain}: the proof of Lemma~\ref{lemma:dbn_explain}
    \item Appendix~\ref{appendix:proof:inexplanability}: the proof of Lemma~\ref{lemma:inexplanability}
    \item Appendix~\ref{appendix:full_edge}: the Unidentifiable Proof for Edge-perturbation  (Sect.~\ref{sect:edge}).
    \item Appendix~\ref{appendix:nodeedge:forwarding}: the analysis of the forwarding computations of a model constructed in Sect.~\ref{sect:node_edge}.
    \item Appendix~\ref{appendix:proof:dbn_explain_2}: the proof of Lemma~\ref{lemma:dbn_explain_2}.
    \item Appendix~\ref{appendix:inexplanability_nodeedge}: the proof of Lemma~\ref{lemma:inexplanability_nodeedge}
    \item Appendix~\ref{appendix:theorem_nodeedge}: the proof of Theorem~\ref{theorem:inexplanability_node_edge}.
    \item Appendix~\ref{appendix:UP_for_gnns}: the discussion on our proofs in GNNs.
\end{itemize}

\section{Bayesian Network and Dynamic Bayesian Network} \label{appendix:prelim_bn}

 Bayesian network (BN)~\cite{PEARL198877} is a Probabilistic Graphical model~\cite{Koller2009}, which represents the conditional dependencies among variables via a directed acyclic graph. The edges in a BN provide users important information, i.e. conditional independence claims, about the relationship between variables in examined system. BN may be constructed either manually with knowledge of the underlying domain, or automatically from datasets by appropriate algorithms. Intuitively, a sparse BN implies its variables can be factorized more easily due to  many conditional independence claims in the graph. Because of this intuition, BN has been used to explain predictions made by GNNs~\cite{minhpgm}. Readers can refer to \cite{Koller2009} for a good overview of BNs.

 Dynamic Bayesian Network (DBN)~\cite{Dubois1992} can be considered as an extension of BN where temporal information is integrated via edges among variables in adjacent time steps. Denote $\{X^{(t)} \}_{t=0}^T$ a set of random vectors with time index $t$, a DBN is a BN modelling those variables and is specified by:
 \begin{itemize}
     \item A BN $\mathcal{B}_0$ consists of variables in $ X^{(t=0)}$ and the corresponding probability distribution on those variables, i.e. $P_0(X^{(t=0)})$.
     \item A set of transition BN $\mathcal{B}_\rightarrow^t$ contains variables in $ X^{(t)} \cup  X^{(t+1)}$. $\mathcal{B}_\rightarrow^t$ is equipped with a probability distribution determining the probability of variables in $t+1$ given variables in $t$, i.e. $P^t_\rightarrow(X^{(t+1)} | X^{(t)})$.
     \item The DBN consists of $\mathcal{B}_0$, $\mathcal{B}_\rightarrow^t$ and the corresponding probability distributions.
 \end{itemize}
 The joint distribution of variables on the DBN then can be factorized by:
 \begin{align*}
     P\left(X^{(0)}, \cdots, X^{(T)} \right)  =
     P_0\left(X^{(0)}\right) \prod_{t=0}^{T-1} P_\rightarrow^t\left(X^{(t+1)}|X^{(t)}\right)
 \end{align*}
 
\begin{figure}[ht]
\centering
\begin{subfigure}{.3\linewidth}
  \centering
  \includegraphics[height=0.8\linewidth]{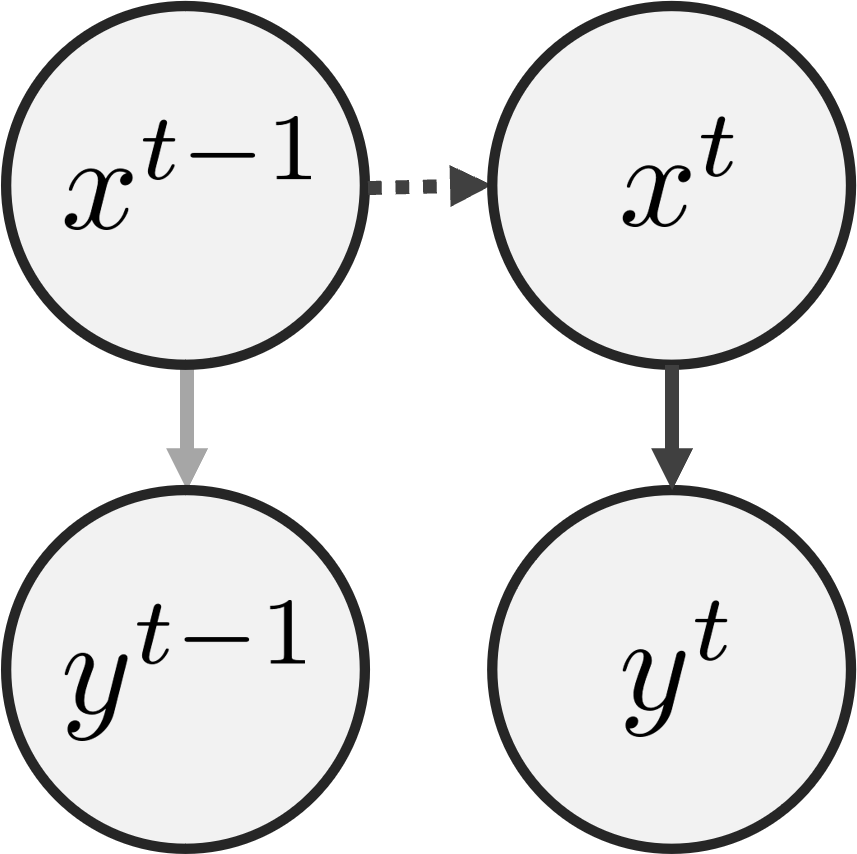}
  \caption{A 2TBN.}
        \label{fig:roll}
\end{subfigure}%
\begin{subfigure}{.6\linewidth}
  \centering
  \includegraphics[height=0.4\linewidth]{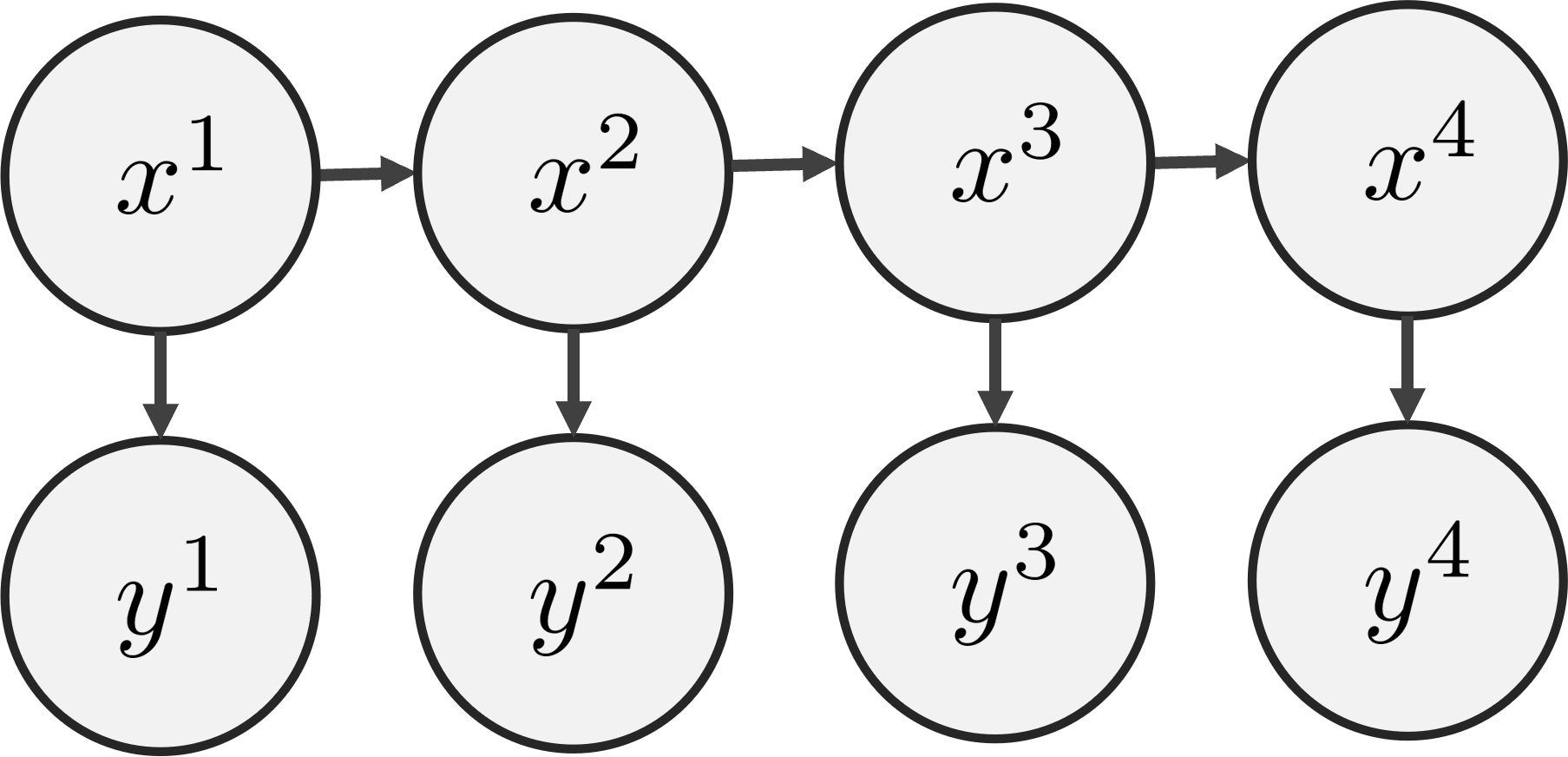}
  \caption{The unroll BNs.}
        \label{fig:unroll}
\end{subfigure}
\caption{An example of the 2TBN and its unroll BN. The intra-slice connections are in solid-lines and the inter-slice connections are in dashed-lines. The brighter solid-line can be omitted as it can be inferred from other edges. }
\label{fig:intro_dbns}
\end{figure}

  When the transition probabilities are the same for all time step, i.e. $P^t\rightarrow$ are independent of $t$, we have one of the most common DBN, which is the Two-Timeslice Bayesian Network (2TBN)~\cite{Murphy2002DynamicBN}. The 2TBN then can be modeled using only 2 BNs, the first contains the prior probability distribution and the second models the transition, which is the form of all DBNs that we illustrated in our proofs. That reference also shows that any systems with dependency among longer temporal windows, i.e. large timeslice, can be reduced to 2TBNs by adjusting the variables.  Figs.~\ref{fig:roll} and \ref{fig:unroll} show an example 2TBN and its unrolled 4 time-step BN. Readers can find more details of DBNs and how to learn them in \cite{Murphy2002DynamicBN, Pamfil2020DYNOTEARSSL}.
  
\section{Forwarding computation of $\Phi_1^v$ and $\Phi_2^v$} \label{appendix:node:forwarding}



Since $h_t, h_s$, and $h_z$ are described in the main manuscript, the signals requires further explanation are $ho^+_i$ and $ho^-_i$. It is easier to understand those two signals by considering their sum, which is the difference between $a_i^{(t, l)}$ and $ht_i^{(t, l-1)} = H^{(t-1)}_i$. Thus, the maximum signal that node $i$ has received can be computed by the following sum:
\begin{align}
    &\frac{1}{2} \left( hr_i^{(t,L)} + ht_i^{(t,L)}+ho_i^{+(t,L)}+ho_i^{-(t,L)}  \right) \nonumber \\
    = & \frac{1}{2} \left( a_i^{(t,L)} + H_i^{(t-1)}+\left| a_i^{(t, l)} - H_i^{(t-1)}\right|  \right) \nonumber \\
    = & \max \left \{ a_i^{(t,L)} , H_i^{(t-1)} \right\} = \max \left \{ hr_i^{(t,L)} , H_i^{(t-1)} \right\} \label{eq:summax}
\end{align}
By setting the READOUT as in Eq.~\ref{eq:endreadout}, we can zero-out the output's signal of nodes with large  $hz_i$. In summary, from the setting shown in Table~\ref{table:h0}, we have:
\begin{align}
    H_i^{(t)} =\left\{\begin{matrix}
    \max \left \{ hr_i^{(t,L)} , H_i^{(t-1)} \right\} 
     &\textup{for } i =1
\\ 
    0  &\textup{for } i \in \{2,3,4\} \label{eq:max}
\end{matrix}\right.
\end{align}
As the prediction of the model $Y$ is set to $ H^{(t=2)}$, to  show that the models indeed compute the max signal of node $i=3$ and output it at node $i=1$, we just need to verify that $hr_1^{(t,L)} = X_3^{(t)}$ for all $t$. 
We now show the claim for $\Phi_1^v$. 

For $l=1$, we have $m_{32}^{(t, 1)} =   hr_3^{(t, 0)} = X_3^{(t)}$ from Eq.~\ref{eq:m1}. Thus, $hr_{2}^{(t,1)} = a_{2}^{(t, 1)} = m_{32}^{(t, 1)} =   X_3^{(t)}$ (Eq.~\ref{eq:a1} and \ref{eq:update}). 

Similarly, for $l=2$, $m_{21}^{(t, 2)} =   hr_2^{(t, 1)} = X_3^{(t)}$ and $hr_{1}^{(t,2)} = a_{1}^{(t, 2)} = m_{21}^{(t, 2)} =  X_3^{(t)}$, which shows that the model fulfills the training task. Note that we have $ a_{2}^{(t, 1)} = m_{32}^{(t, 1)}$  and $ a_{1}^{(t, 2)} = m_{21}^{(t, 2)}$  because there is no message sending from node 1 and node 4. The actual values for all $hr_i$ during the forwarding computations are shown in Table~\ref{table:forward}. 

\begin{table}[ht]
\centering
\caption{Hidden features of the TGNN $\Phi_1^v$ for input $ X^{(t=1)} = [\alpha_1,\alpha_2,\alpha_3,\alpha_4]$ and $X^{(t=2)} = [\beta_1,\beta_2,\beta_3,\beta_4]$. All values are less than $\min \{k_s, k_z \}$.} 
\label{table:forward}
\scalebox{0.77}{
\begin{tabular}{|c|c|c|c|c|}
\hline
{Variable} & $H_i$      & $X = hr^{l=0}_i$  & $hr^{l=1}$        & $hr^{l=2}$        \\ \hline
$t=1$             & $0,0,0,0$   & $\alpha_1,\alpha_2,\alpha_3,\alpha_4$ & $\alpha_2,\alpha_3,\alpha_2,\alpha_3$ & $\alpha_3,\alpha_2,\alpha_3,\alpha_2$ \\ \hline
$t=2$             & $\alpha_3,0,0,0$ & $\beta_1,\beta_2,\beta_3,\beta_4$ & $\beta_2,\beta_3,\beta_2,\beta_3$ & $\beta_3,\beta_2,\beta_3,\beta_2$ \\ \hline
\end{tabular}
}
\end{table}

The proof for $\Phi_2^v$ trivially follows as all arguments hold by swapping node $2$ with node $4$.

\section{Proof of Lemma~\ref{lemma:dbn_explain}} \label{appendix:proof:dbn_explain}
 
 \textbf{Lemma 1.} \textit{The DBN $\mathcal{B}_1^v$ ($\mathcal{B}_2^v$) in Fig.~\ref{fig:explanations} can embed all information of the hidden features of TGNN $\Phi_1^v$ ($\Phi_2^v$ ) without any loss when the input signal is bounded by $K \vcentcolon = \min \{k_s, k_z\}$. Furthermore, the DBN is a minimal.}


\begin{proof}
To show that the DBN  $\mathcal{B}_1^v$ can represent the TGNN $\Phi_1^v$  without any loss, we show:
\begin{itemize}
    \item  $\mathcal{B}_1^v$ can express how the predictions $H_i^{(t)}$ are generated.
    \item  $\mathcal{B}_1^v$ can express how the messages propagate in the model.
\end{itemize}

Showing the first bullet is easier since we only need to consider $i=1$ as $H_i^{(t)} = 0$ for all other nodes. Because $H_1^{(t)} = \max \left \{ X_3^{(t)} , H_1^{(t-1)} \right\} $ (shown below Eq.~\ref{eq:max}), a path from $\mathcal{V}_3^t$ to $\mathcal{V}_1^t$ and a path from $\mathcal{V}_1^{t-1}$ to $\mathcal{V}_1^t$ are sufficient to represent how the predictions are generated. For the second bullet, from Table~\ref{table:h0}, as long as the inputs are bounded by $K$, only nodes 2 and 3 are sending out messages. It is also clear from Eq.~\ref{eq:m1}, \ref{eq:a1} and \ref{eq:update} that:
\begin{align*}
    &X_2^{(t)} \rightarrow m_{21}^{(t,1)} \rightarrow a_1^{(t,1)} \rightarrow hr_1^{(t,1)} \rightarrow \varnothing \\
    &X_2^{(t)} \rightarrow m_{23}^{(t,1)} \rightarrow a_3^{(t,1)} \rightarrow hr_3^{(t,1)}  \rightarrow m_{32}^{(t,2)} \textup{ and } m_{34}^{(t,2)} \\
    &X_3^{(t)} \rightarrow m_{32}^{(t,1)} \rightarrow a_2^{(t,1)} \rightarrow hr_2^{(t,1)}  \rightarrow m_{23}^{(t,2)} \textup{ and } m_{21}^{(t,2)}\\
    &X_3^{(t)} \rightarrow m_{34}^{(t,1)} \rightarrow a_4^{(t,1)} \rightarrow hr_4^{(t,1)}  \rightarrow \varnothing
\end{align*}
where the arrow means \textit{determining} and the notation $\rightarrow \varnothing $ means the signals result in no other messages. From the chains, we see that the messages sending from node 2 and 3 are only dependent on the other's. Hence, this dependency can be captured by an edge between $\mathcal{V}_2^t$ and $\mathcal{V}_3^t$.

The above arguments also show that $\mathcal{B}_1^v$ is minimal. Specifically, the edges $(\mathcal{V}_1^{t-1}, \mathcal{V}_1^t)$ and $(\mathcal{V}_2^t$, $\mathcal{V}_3^t)$ are necessary because of the temporal dependency $H_1^{(t-1)} \rightarrow H_1^{(t)} $ and the messages' dependency between nodes 2 and 3. We then require a path from $\mathcal{V}_3^t$ to $\mathcal{V}_1^t$ to capture the dependency $X_3^{(t)} \rightarrow H_1^{(t)}$ when $X_3^{(t)} >H_1^{(t-1)} $. Thus, at least another edge is needed. Since $\mathcal{B}_1^v$ has 3 edges, it is minimal.
\end{proof}

\section{Proof of Lemma~\ref{lemma:inexplanability}} \label{appendix:proof:inexplanability}

\textbf{Lemma 2.}
\textit{For all $ \bm X $ such that $X_i^{(t)} \leq \min \{k_s, k_z\}$:}
\begin{align*}
    \Phi_1^v(\bm X) =  \Phi_2^v(\bm X)
\end{align*}

\begin{proof}
From the examination of the forwarding computation (Appx.~\ref{appendix:node:forwarding}), we know that both models satisfy Eq.~\ref{eq:task_node} for all $\bm X$ bounded by $\min \{k_s, k_z\}$. Thus, their outputs on those $\bm X$ are the same. Therefore, we have the Lemma.
\end{proof}




\section{Unidentifiable Proof for Edge-perturbation}  \label{appendix:full_edge}

This section is about the Unidentifiable Proof for Edge-perturbation class $\mathcal{G}_e$. As stated in the main manuscript, we will show that removing edges from the  input graph is not enough to identify all nodes contributing to a max operation conducted by the TGNNs.

\begin{figure}[ht]
\centering
\begin{subfigure}{.48\linewidth}
  \centering
  \includegraphics[height=0.49\linewidth]{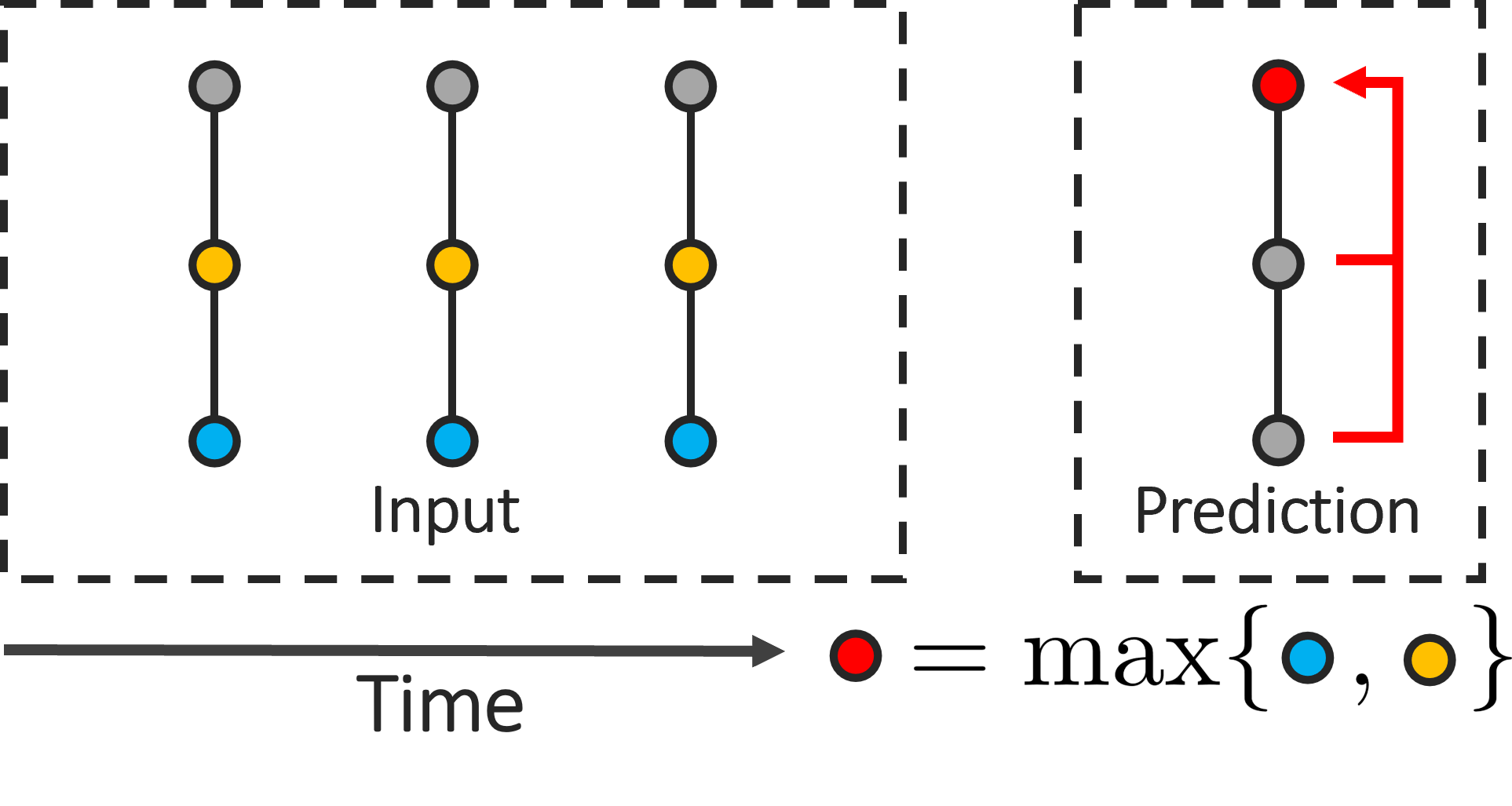}
  \caption{The training task.}
        \label{fig:task_edge_botbot}
\end{subfigure}%
\begin{subfigure}{.51\linewidth}
  \centering
  \includegraphics[height=0.44\linewidth]{LaTeX/figures/DBNes.png}
  \caption{The DBNs.}
        \label{fig:dbn_edge_botbot}
\end{subfigure}
\caption{The components for $\mathcal{G}_e$'s Unidentifiable Proof.}
\label{fig:Ge_components}
\end{figure}

\textbf{The Training Task.} Our proof considers a graph of 3 nodes forming a line. The task (Fig.~\ref{fig:task_edge_top}, redrawn in Fig.~\ref{fig:task_edge_botbot}) is to recognize the maximum positive inputs observed in node 2 and 3, and return result at node 1:
\begin{align}
 Y_1^{(t)} &= \max \left\{0, X_2^{(t')} \textup{ and } X_3^{(t')} , 0\leq t' \leq t \right\} \label{eq:task_edge_appx}
\end{align}
The outputs on other nodes are zeros.

It is clear that the DBN $\mathcal{B}_1^e$ (Fig.~\ref{fig:dbn_edge}, redrawn in Fig.~\ref{fig:dbn_edge_botbot}) can fulfill this training task: from edge $\mathcal{V}_{3}^{(t)}-\mathcal{V}_{2}^{(t)}$, $\mathcal{V}_{2}^{(t)}$ can be set to be the maximum of those 2 variables. That maximum is then transmitted to $\mathcal{V}_{1}^{(t)}$ through $\mathcal{V}_{2}^{(t)}-\mathcal{V}_{1}^{(t)}$. Finally, the edge $\mathcal{V}_{1}^{(t-1)}-\mathcal{V}_{1}^{(t)}$ helps the model determine the maximum signal node 1 has ever received. On the other hand, the DBN $\mathcal{B}_2^e$ (Fig.~\ref{fig:dbn_edge}, redrawn in Fig.~\ref{fig:task_edge_botbot}) disconnects node 3 and obviously cannot achieve the task. However, we will construct two models whose Transparent Models are $\mathcal{B}_1^e$ and $\mathcal{B}_2^e$ and all Edge-perturbation explanation methods cannot differentiate them.


\textbf{The Models.} We use the same architecture with as in Sect.~\ref{sect:node} (Fig.~\ref{fig:model}). 
The hidden feature vector of each node $i \in \{1,2,3\}$ have 5 main features, i.e. $\bm h_i  = [hr_i, ht_i, hs_i, hz_i, hl_i]$, and 6 additional features just for output purposes, denoted by $\bm {ha}_i  = [hrl^+_i, hrl^-_i,hrt^+_i, hrt^-_i,hlt^+_i, hlt^-_i ]$. Similar to the proof for Node-perturbation, $\bm h_i^{(t,l=0)}$ refers to the model's input:
\begin{align}
    \bm h_i^{(t,l=0)} &= \left[X_i^{(t)}, H_i^{(t-1)}, *, *, 0\right]
    \label{eq:u0_3}
\end{align}

The only new feature in $\bm h_i$ is $hl_i$, which is constructed to be the lag version of $hr_i$ by setting the UPD:
\begin{align}
         hl_i^{(t,l)} =&  \textup{ ReLU} \left( hr_i^{(t, l-1)} \right) \label{eq:lag}
\end{align}
$hl_i$ is zero at $l=0$ because there is no previous lag message. Beside from that, all features in $\bm h_i$ have the same meaning as described in Sect.~\ref{sect:node}. Since $hs_i$ and $hz_i$ are still constant features, we need to specify their values, which are provided in Table~\ref{table:h1}. Furthermore, the models have the same MSG, AGG and UPD functions as described from Eq.~\ref{eq:m1} to  Eq.~\ref{eq:uend}. It is noteworthy to point out that the only difference between the two models is in node 3: while it can send message in $\Phi_1^e$ (as $hs_3$  in $\Phi_1^e$ is 0), it cannot in $\Phi_2^e$ (as $hs_3 = k_s$ in $\Phi_2^e$).

\begin{table}[ht]
\centering
\caption{The constant features of the TGNNs in $G_e$'s proof.}
\label{table:h1}
\scalebox{0.8}{
\begin{tabular}{|l|cc|cc|cc|}\hline
\textbf{Node}            & \multicolumn{2}{c|}{\textbf{1}}      & \multicolumn{2}{c|}{\textbf{2}}      & \multicolumn{2}{c|}{\textbf{3}}          \\ \hline
\textbf{Hidden features} & \multicolumn{1}{c|}{$hs_1$} & \multicolumn{1}{c|}{$hz_1$} &  
                           \multicolumn{1}{c|}{$hs_2$} & \multicolumn{1}{c|}{$hz_2$} & 
                           \multicolumn{1}{c|}{$hs_3$} & \multicolumn{1}{c|}{$hz_3$}    \\ \hline
\textbf{TGNN $\Phi_1^e$}         & \multicolumn{1}{c|}{$k_s$}      & \multicolumn{1}{c|}{0}      & 
                           \multicolumn{1}{c|}{0}      & \multicolumn{1}{c|}{$k_z$}       & 
                           \multicolumn{1}{c|}{0}      & \multicolumn{1}{c|}{$k_z$}    \\ \hline
\textbf{TGNN $\Phi_2^e$}         & \multicolumn{1}{c|}{$k_s$}      & \multicolumn{1}{c|}{0}      & 
                           \multicolumn{1}{c|}{0}      & \multicolumn{1}{c|}{$k_z$}    & 
                           \multicolumn{1}{c|}{$k_s$}      & \multicolumn{1}{c|}{$k_z$}    \\ \hline
\end{tabular}
}
\end{table}

For the 6 additional features $\bm {ha}_i$, they are zeros at initialization. Their update are set as:
\begin{align*}
         hrl_i^{+(t,l)} =&  \textup{ReLU} \left(  a_i^{(t, l)} -   hr_i^{(t,l-1)} \right)\\
         hrl_i^{-(t,l)} =&  \textup{ReLU} \left(  hr_i^{(t,l-1)}-  a_i^{(t, l)}\right)\\
         hrt_i^{+(t,l)} =&  \textup{ReLU} \left( a_i^{(t, l)} - ht_i^{(t,l-1)}  \right)\\
         hrt_i^{-(t,l)} =&  \textup{ReLU} \left( ht_i^{(t,l-1)}  - a_i^{(t, l)}\right)\\
         hlt_i^{+(t,l)} =&  \textup{ReLU} \left( hr_i^{(t,l-1)}- ht_i^{(t,l-1)} \right)\\
         hlt_i^{-(t,l)} =&  \textup{ReLU} \left(ht_i^{(t,l-1)} - hr_i^{(t,l-1)}\right)
\end{align*}
It is clear that the sum of features in $\bm {ha}_i$ is the sum of $| a_i^{(t, l)} - hr_i^{(t, l-1)}|$, $| a_i^{(t, l)} - ht_i^{(t, l-1)}|$ and $| hr_i^{(t, l-1)} - ht_i^{(t, l-1)}|$. Since $ht_i^{(t,l)} = ht_i^{(t,l-1)}$, $hr_i^{(t,l)} = a_i^{(t, l)}$ and $hl_i^{(t,l)} = hr_i^{(t,l-1)}$, we have the following expression:
\begin{align*}
    \tau_i^{(t,l)} \vcentcolon= \frac{ \bm 1^\top \bm {ha}^{(t,l)}_i + hr_i^{(t,l)} + hl_i^{(t,l)} + ht_i^{(t,l)}}{3}
\end{align*}
is the maximum of $ hr_i^{(t,l)}$, $ hl_i^{(t,l)}$ and $ht_i^{(t,l)}$. With that, we set the READOUT and the prediction as:
\begin{align}
    H_i^{(t)} &= \textup{ReLU} \left( \tau_i^{(t,l=2)} - hz_i^{(t,l=2)}\right) \\
    Y &= H^{(t=2)}  \label{eq:endreadoutz} 
\end{align}

The construction of $\Phi_1^e$ is only different to that of $\Phi_1^v$ (Sect.~\ref{sect:node}) in $hl_i$, $\bm {ha}_i$ and the READOUT. The first 4 features in $\bm h_i$ in $\Phi_1^e$ behave exactly the same as those in $\Phi_1^v$ on node $i \in \{1,2,3\}$. As we have shown $hr_1^{(t,l=2)} = X_3^{(t)}$ and $hr_1^{(t,l=1)}= X_2^{(t)}$ in  $\Phi_1^v$ (Appx.~\ref{appendix:node:forwarding}), it follows that  $hr_1^{(t,l=2)}= X_3^{(t)}$ and $hl_1^{(t,l=2)}= X_2^{(t)}$ in $\Phi_1^e$  as $hl_1$ is the lag-1 version of $hr_1$ due to Eq.~\ref{eq:lag}. By setting the READOUT as Eq.~\ref{eq:endreadoutz}, similar analysis as shown at Eq.~\ref{eq:max} gives us the temporal output of the model is:
\begin{align*}
    H_1^{(t)} &=
    \max \left \{ hr_1^{(t,L)}, hl_1^{(t,L)} , H_1^{(t-1)} \right\} \\
    & = \max \left \{  X_3^{(t)},  X_2^{(t)} , H_1^{(t-1)} \right\}
\end{align*}
and $ H_i^{(t)} = 0$ for $i \in \{2, 3\}$. This means the model's output satisfies the training task  (Eq.~\ref{eq:task_edge}, which is restated in Eq.~\ref{eq:task_edge_appx}).

By setting $hs_3 = k_s$ in $\Phi_2^e$, we prevent node 3 from sending messages to node 2. This makes $hr_1^{(t,l=2)}= 0$ as there is no message coming to node 1 at $l=2$. This means the output of $\Phi_2^e$ at node 1 is:
\begin{align*}
    H_1^{(t)} =
    \max \left \{ hl_1^{(t,L)} , H_1^{(t-1)} \right\} =  \max \left \{  X_2^{(t)} , H_1^{(t-1)} \right\}
\end{align*}
Thus, the output of $\Phi_2^e$ is as described in Eq.~\ref{eq:task_edge_2}.

By simply comparing Eq.~\ref{eq:task_edge} with Eq.~\ref{eq:task_edge_2}, we can see that $\Phi_1^e$ and  $\Phi_2^e$ are different. However, for inputs with $X_2^{(t)} > X_3^{(t)}$, the responses of the two models are the same, which is stated in Lemma.~\ref{lemma:inexplanability_edge}. Below is the restatement of the Lemma and its proof:

\textbf{Lemma 3.} 
 \textit{ For the task in Fig~\ref{fig:task_edge_top}, denote $\bar{A}$ the adjacency matrix obtained by either keeping the input adjacency matrix $A$ unchanged or by removing some edges. For any given $\bm X $ such that $X_i^{(t)} \leq \min \{k_s, k_z\}$ and $X_2^{(t)} > X_3^{(t)}$, we have}
 \begin{align*}
    \Phi_1^e(\bm X, \bar{A}) =  \Phi_2^e(\bm X, \bar{A}) 
\end{align*}

\begin{proof}
We only need to consider the models' outputs at node 1 since outputs of all other nodes are 0 (as $hz_i = k_z$ for $i \in \{2,3\}$, given in Table~\ref{table:h1}). 
If no edge is removed, from the analysis of the forwarding computation (below Eq.~\ref{eq:endreadoutz}), we know that both models return the maximum of $X^{(t)}_2$ at node 1 as $X^{(t)}_2> X^{(t)}_3$.
If the edge between node 1 and node 2 is removed, there is no message coming to node 1 and $hr_1^{(t,l=2)}$ in both models will be 0.
The remained case is when only the edge between node 2 and 3 is removed. In this case, $\Phi_1^e$ simply becomes $\Phi_2^e$ and their outputs must be the same. We then have the Lemma.
\end{proof}

\textbf{The Transparent Models of $\Phi_1^e$ and $\Phi_2^e$.}
Here, we summarize the arguments in the main manuscript to establish $\mathcal{B}_1^e = \mathcal{I}(\Phi_1^e)$ for $\bm X$ bounded by $\min \{k_s, k_z\}$, and $\mathcal{B}_2^e = \mathcal{I}(\Phi_2^e)$ for $\bm X$ bounded by $\min \{k_s, k_z\}$ and $X_2^{(t)} > X_3^{(t)}$. The first claim is because  $\Phi_1^e$ is almost identical to $\Phi_1^v$ (Sect.~\ref{sect:node}) without node 4. The second is because $\Phi_2^e$ is $\Phi_1^e$ but node 3 is disconnected.

\textbf{Unidentifiable Proof.} 
We are now ready to proof the Unidentifiable result fo Edge-perturbation. We restate Theorem~\ref{theorem:inexplanability_node_edge} and provide its proof in the following:

\textbf{Theorem 2.} \textit{For a TGNN $\Phi$ (Eq.~\ref{eq:tgnnlast}), denote $\mathcal{P} \vcentcolon= \{ ( X, \bar{A}, \Phi(X, \bar{A})) | X_i \leq K\}_{\bar{A}} $, i.e. the set of Edge-perturbation-response of $\Phi$ where $\bm X$ are fixed and bounded by $K$, and $\bar{A}$ is defined as in Lemma~\ref{lemma:inexplanability_edge}. Denote $g$ an arbitrary algorithm accepting $\mathcal{P}$ as inputs.  For any $K>0$ and any $g$, there exists a $\Phi$ such that:}
\begin{enumerate}
    \item \textit{For the interpretable domain of DBNs, the Transparent Model of $\Phi$ exists for all inputs in $\mathcal{P}$.}
    \item \textit{$g$ cannot determine the Transparent Model of $\Phi$.}
\end{enumerate}

\begin{proof}
We first choose $k_s$ and $k_z$ in Table~\ref{table:h1} to $K$ in the Theorem. We then construct $\Phi_1^e$ and $\Phi_2^e$ as described in Sect.~\ref{sect:edge}. Denote $\mathcal{P}_1$ and $\mathcal{P}_2$ the sets of Edge-perturbation-response of $\Phi_1^e$ and $\Phi_2^e$, respectively. Note that, from the discussion of Transparent Models, we have $\mathcal{B}_1^e$ and $\mathcal{B}_2^e$ are the Transparent Models the two models.

Given an Edge-perturbation-response, suppose $g$ returns either $\mathcal{B}_1^e$ or $\mathcal{B}_2^e$. Due to Lemma~\ref{lemma:inexplanability_edge}, $\mathcal{P}_1$ is the same as $\mathcal{P}_2$; therefore, the outputs of $g$ on the 2 perturbation-response sets must be the same. Hence, similar arguments as in the proof of Theorem~\ref{theorem:inexplanability_node} (shown in the main manuscript) give us Theorem~\ref{theorem:inexplanability_edge}.
\end{proof}

\section{Forwarding computation of $\Phi_2^a$} \label{appendix:nodeedge:forwarding}

To analyze the forwarding computation of $\Phi_2^a$, we have the following observations:
\begin{itemize}
    \item The message $m_{ij}$ (specified at Eq.~\ref{eq:mdev}) is simply the maximum of $hr_i$ and $ht_i$ (similar to the analysis at Eq.~\ref{eq:summax}, Appx.~\ref{appendix:node:forwarding}) when $hs_i = 0$.
    \item Because of the READOUT (specified below Eq.~\ref{eq:mdev}), $H_2^{(2)}$ and $ht_2$ are always zeros. Thus, $m_{2j}$ is always $hr_2$.
    \item Since node 2 only receives message from node 3 (as $hs_1=0$) and the input message, $hr_2^{(t,l)}$ is always the message sent from 3 for $l$ odd. Combining with the above point,  $m_{21}^{(t,l=2)}$ is always $m_{32}^{(t,l=1)}$.
     \item As node 1 only connects to node 2, $hr_1^{(t,l=2)}$ is always $m_{21}^{(t,l=2)}$ , which is $m_{32}^{(t,l=1)}$.
\end{itemize}
From those observations, we have $hr_1^{(t,l=2)}$ is the maximum of $hr_3^{(t,l=0)}$ and $ht_3$. If $ht_3$ is always the maximum input in the past and $hr_3^{(t,l=0)}$ is the current input of node 3, then we have the model fulfill the training task (Fig.~\ref{fig:task_node_edge_top}). 

To see that $ht_3$ is always the maximum input in the past, we refer to Table~\ref{table:forward2} tracking the received message $hr_i$ and the out-going message $m_{i*}$ for arbitrary inputs $ X^{(t=1)} = [\alpha_1,\alpha_2,\alpha_3]$ and $X^{(t=2)} = [\beta_1,\beta_2,\beta_3]$. The claim for the input's length $T=2$ is trivial: for the first time-step, the values are the same as in the case in Table~\ref{table:forward} since $ht_i,ho_i^+$ and $ho_i^-$ are zeros and the sending message in Eq.~\ref{eq:mdev} is just $hr_i$ as before. For $t=2$, we have $ht_3 = \alpha_3$ (because only $hz_3 = 0$), which is indeed the maximum signal in the past of node 3. For larger time-step $T$, we can further examine the Table~\ref{table:forward2} and deduce that claim: at $l=0$, node $3$ sends out $\max \{ X_3^{(t)}, H_3^{(t-1)}\}$. At $l =1$, this message is received at node 2. Finally, this message is sent back to node 3 at $l=2$.

\begin{table}[ht]
\centering
\caption{Hidden features of the TGNN $\Phi_2^a$  for input $ X^{(t=1)} = [\alpha_1,\alpha_2,\alpha_3]$ and $X^{(t=2)} = [\beta_1,\beta_2,\beta_3]$. $\gamma_3 = \max \{\alpha_3, \beta_3\}$ and $\gamma_2 = \max \{\gamma_3, \beta_2\}$}
\label{table:forward2}
\scalebox{0.82}{
\begin{tabular}{|cc|c|c|c|}
\hline
\multicolumn{2}{|c|}{Variable}                                                                                                 & Input                        & Layer 1                      & Layer 2                      \\ \hline
\multicolumn{1}{|c|}{\multirow{2}{*}{\begin{tabular}[c]{@{}c@{}}$t=1$\\ $H^{(t)}=[0,0,0]$\end{tabular}}}        & $m_{i*}$ &                              & $0,\alpha_2,\alpha_3$        & $0,\alpha_3,\alpha_2$        \\ \cline{2-5} 
\multicolumn{1}{|c|}{}                                                                                    & $hr_i$               & $\alpha_1,\alpha_2,\alpha_3$ & $\alpha_2,\alpha_3,\alpha_2$ & $\alpha_3,\alpha_2,\alpha_3$ \\ \hline
\multicolumn{1}{|c|}{\multirow{2}{*}{\begin{tabular}[c]{@{}c@{}}$t=2$\\ $H^{(t)}=[0,0,\alpha_3]$\end{tabular}}} & $m_{i*}$ &                              & $0,\beta_2,\gamma_3$         & $0,\gamma_3,\gamma_2$        \\ \cline{2-5} 
\multicolumn{1}{|c|}{}                                                                                    & $hr_i$               & $\beta_1,\beta_2,\beta_3$    & $\beta_2,\gamma_3,\beta_2$   & $\gamma_3,\gamma_2,\gamma_3$ \\ \hline
\end{tabular}
}
\end{table}

\section{Proof of Lemma~\ref{lemma:dbn_explain_2}} \label{appendix:proof:dbn_explain_2}


 \textbf{Lemma 4.} \textit{The DBN $\mathcal{B}_2^a$ in Fig.~\ref{fig:dbn_node_edge} can embed all information of the hidden features of TGNN $\Phi_2^a$  without any loss when the input signal is bounded by $K \vcentcolon = \min \{k_s, k_z\}$. Furthermore, the DBN is a minimal.}

\begin{proof}
The proof has the same structure as in Lemma~\ref{lemma:dbn_explain}, in which we show the DBN can express the predictions and the messages. For the prediction, in the forwarding computation, we have shown $hr_1^{(t,l=2)}$ is the maximum of $hr_3^{(t,l=0)}$ and $ht_3$. Thus, the path $\mathcal{V}_3^{t-1}-\mathcal{V}_3^{t}-\mathcal{V}_2^{t}-\mathcal{V}_1^{t}$ is sufficient to express the prediction. For the messages, since only node 2 and 3 are sending out messages, we only need the edge $\mathcal{V}_3^{t}-\mathcal{V}_2^{t}$ and edge $\mathcal{V}_2^{t}-\mathcal{V}_1^{t}$ to represent them. We can also track the signal via Table~\ref{table:forward2} to verify this. 

It is easy to verify that the DBN $\mathcal{B}_2^a$ is indeed minimal: simply from the fact that the prediction at node 1 depends on the past signal $ht_3$, which means me must maintain a path between $\mathcal{V}_3^{t-1}$ and $\mathcal{V}_1^{t}$. Thus, we cannot remove any edges from  $\mathcal{B}_2^a$ while keeping it consistent with $\Phi_2^a$.
\end{proof}

\section{Proof of Lemma~\ref{lemma:inexplanability_nodeedge}} \label{appendix:inexplanability_nodeedge}

\textbf{Lemma 5.} \textit{For the training task in Fig~\ref{fig:task_node_edge_top}, denote $\bar{A}$ the adjacency matrix obtained by either keeping the input adjacency matrix $A$ unchanged or by removing some edges from $A$.
For all $ \bm X $ such that $X_i^{(t)} \leq \min \{k_s, k_z\}$, we have}
\begin{align*}
    \Phi_1^a(\bm X, \bar{A}) =  \Phi_2^a(\bm X, \bar{A})
\end{align*}

\begin{proof}
First, if $A$ is fixed, from the examination of the forwarding computation, we know that both models satisfy Eq.~\ref{eq:spec} for all $\bm X$ bounded by $\min \{k_s, k_z\}$. Thus, we have the Lemma. 

If the edge between node 1 and 2 is removed, there is no message coming to node 1 and the models' outputs will always be zeros. The remained case is only the edge between node 2 and 3 is removed. In that situation, at $l=1$, there is no incoming message to node 2 (because node 1 does not send message and node 3 is disconnected) and $hr_2^{(t,l=1)} = 0$. This means at $l=2$, there is no incoming message to node 1 because $m_{21}^{(t,l=2)} = hr_2^{(t,l=1)} = 0$. As a result,  the models' outputs will also always be zeros. We then have the Lemma.
\end{proof}

\section{Proof of Theorem~\ref{theorem:inexplanability_node_edge}} \label{appendix:theorem_nodeedge}

\textbf{Theorem 3.}\textit{For a TGNN $\Phi$ (Eq.~\ref{eq:tgnnlast}), denote $\mathcal{P} \vcentcolon= \{ ( X, \bar{A}, \Phi(X, \bar{A}) ) | X_i \leq K\}_{X \in \mathcal{X}, \bar{A}}$, i.e. the set of Node-and-Edge-perturbation-response of $\Phi$ where $\bm X$ are fixed and bounded by $K$, and $\bar{A}$ is defined as in Lemma~\ref{lemma:inexplanability_nodeedge}. Denote $g$ an arbitrary algorithm accepting $\mathcal{P}$ as inputs.  For any $K>0$ and any $g$, there exists a $\Phi$ such that:}
\begin{enumerate}
    \item \textit{For the interpretable domain of DBNs, the Transparent Model of $\Phi$ exists for all inputs in $\mathcal{P}$.}
    \item \textit{$g$ cannot determine the Transparent Model of $\Phi$.}
\end{enumerate}


\begin{proof}
We first choose $k_s$ and $k_z$ in Table~\ref{table:h2} to $K$ in the Theorem. We then construct $\Phi_1^a$ and $\Phi_2^a$ as described in Sect.~\ref{sect:node_edge}. Denote $\mathcal{P}_1$ and $\mathcal{P}_2$ the sets of Node-and-Edge-perturbation-response of $\Phi_1^a$ and $\Phi_2^a$, respectively. Note that, from the discussion of Transparent Models in that section, we have $\mathcal{B}_1^a$ and $\mathcal{B}_2^a$ are the Transparent Models the two models.

Given an Node-and-Edge-perturbation-response, suppose $g$ returns either $\mathcal{B}_1^a$ or $\mathcal{B}_2^a$. From Lemma~\ref{lemma:inexplanability_nodeedge}, $\mathcal{P}_1$ is the same as $\mathcal{P}_2$; therefore, the outputs of $g$ on the 2 perturbation-response sets must be the same. Follow similar arguments as in the proof of Theorem~\ref{theorem:inexplanability_node} (shown in the main manuscript), we have Theorem~\ref{theorem:inexplanability_node_edge}.
\end{proof}

\section{Unidentifiable Proofs for GNNs} \label{appendix:UP_for_gnns}

Note that in base GNNs, we need to drop the temporal dimension in the training task and the interpretable domain. We also use BNs instead of DBNs.

\begin{figure}[ht]
\centering
\begin{subfigure}{.49\linewidth}
  \centering
  \includegraphics[height=0.5\linewidth]{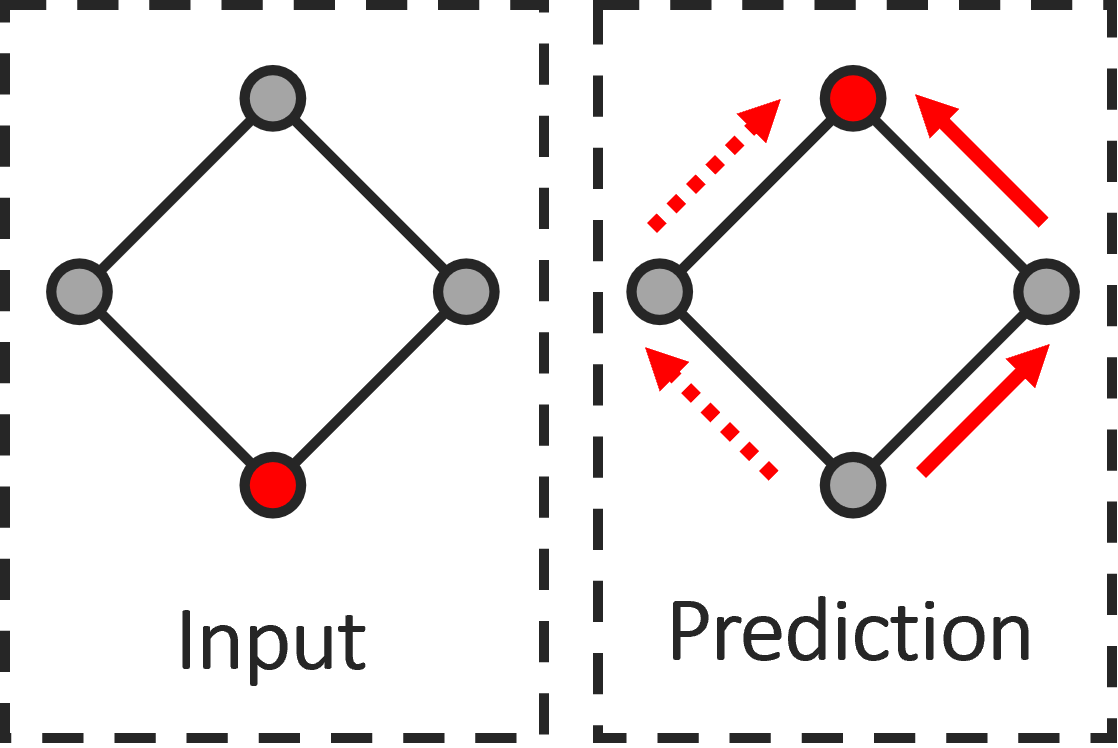}
  \caption{The training task.}
        \label{fig:task_node_gnn}
\end{subfigure}%
\begin{subfigure}{.49\linewidth}
  \centering
  \includegraphics[height=0.48\linewidth]{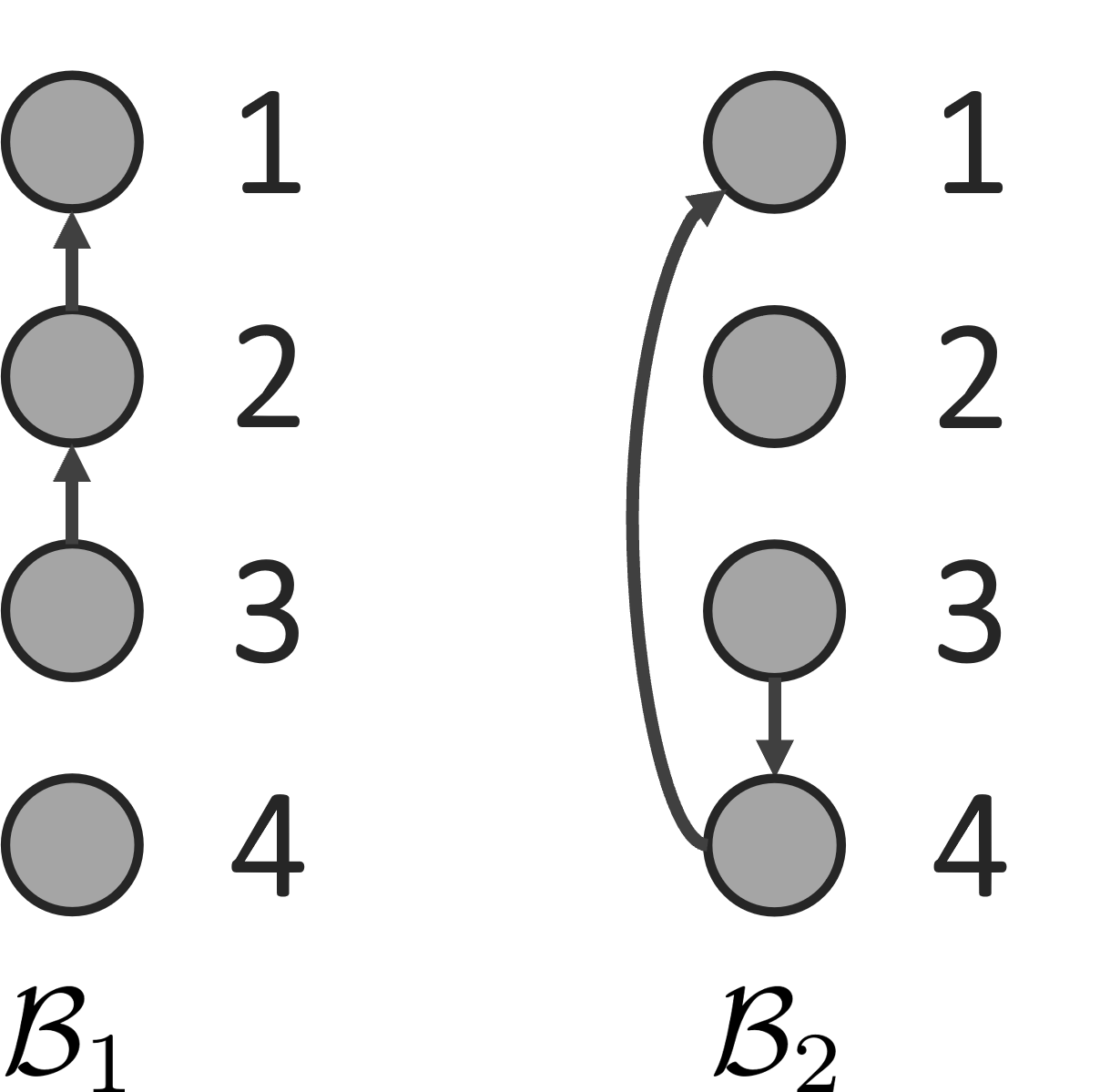}
  \caption{The BNs.}
        \label{fig:dbn_node_gnn}
\end{subfigure}
\caption{The components for Node-perturbation Unidentifiable Proof for base GNNs.}
\label{fig:Gv_gnn}
\end{figure}

\begin{figure}[ht]
\centering
\begin{subfigure}{.49\linewidth}
  \centering
  \includegraphics[height=0.5\linewidth]{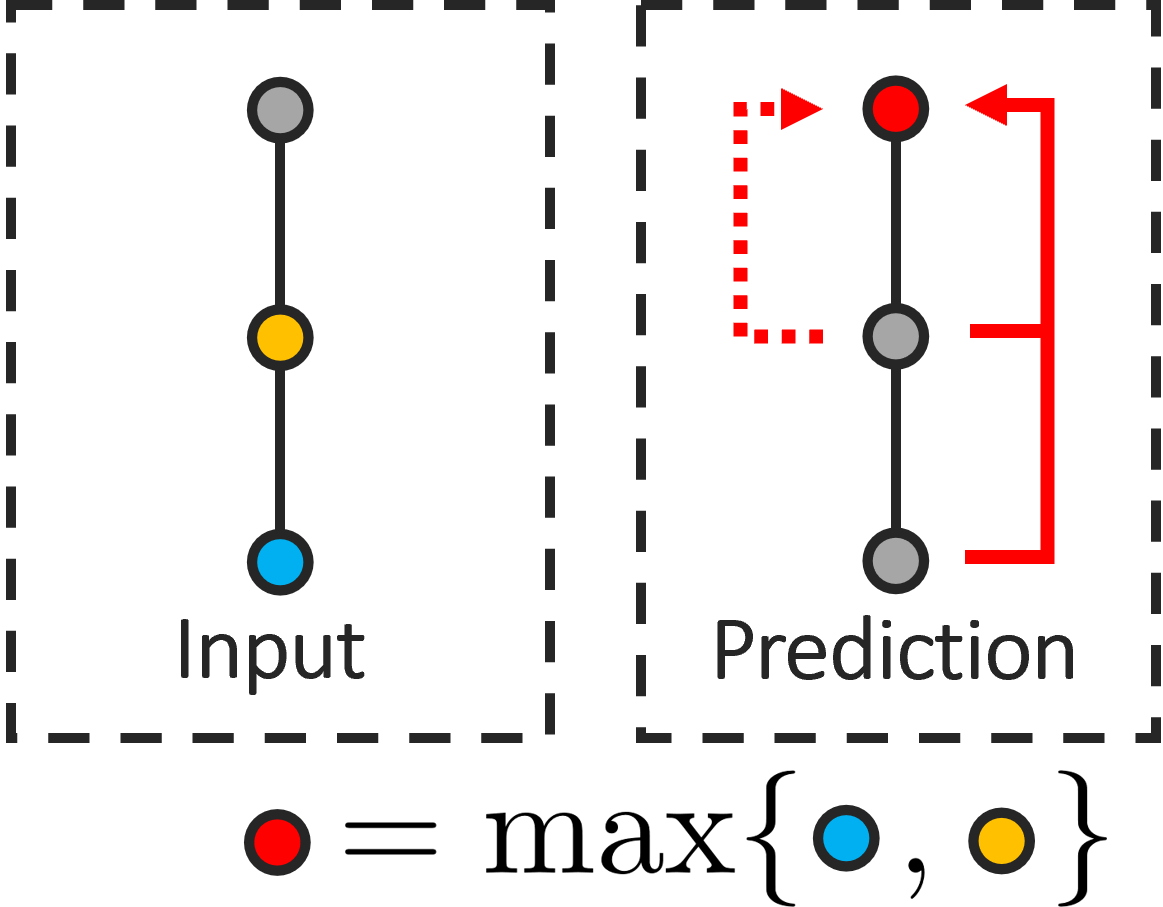}
  \caption{The training task.}
        \label{fig:task_edge_gnn}
\end{subfigure}%
\begin{subfigure}{.49\linewidth}
  \centering
  \includegraphics[height=0.5\linewidth]{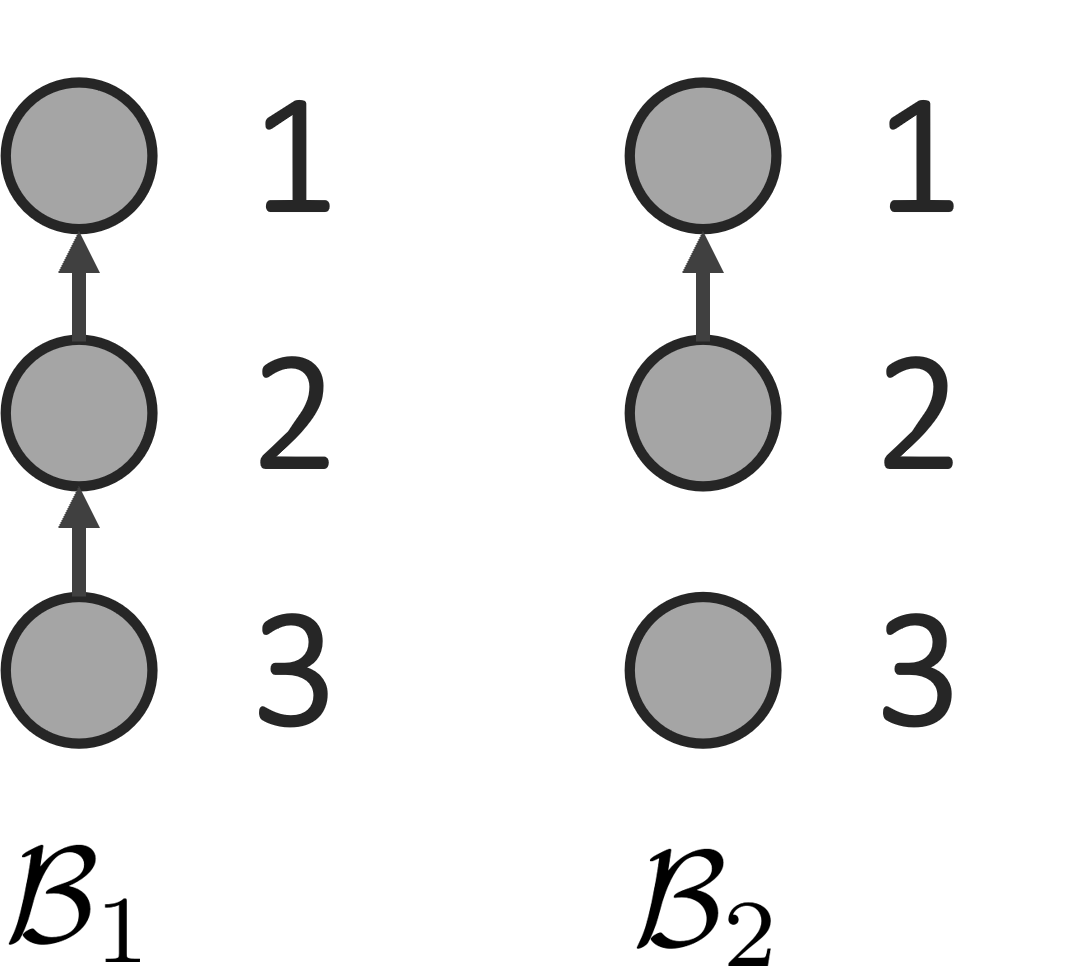}
  \caption{The BNs.}
        \label{fig:dbn_edge_gnn}
\end{subfigure}
\caption{The components for Edge-perturbation Unidentifiable Proof for base GNNs.}
\label{fig:Ge_gnn}
\end{figure}

For the case of Node-perturbation, the new training task will be $Y_1 = X_3$ and all other $Y_i = 0$, which is described in Fig.~\ref{fig:task_node_gnn}. By simply dropping the time index $t$, we can follow the analysis as shown in Sect.~\ref{sect:node} and have the Transparent Models of the two models to be BN $\mathcal{B}_1$ and $\mathcal{B}_2$ shown in Fig.~\ref{fig:dbn_node_gnn}, respectively. As the two models have different Transparent Models but the same responses on all perturbations, we have the Unidentifiable Proof for the base GNNs.

For the Edge-perturbation, the training task will be $Y_1 = \max\{ X_2, X_3 \}$ and all other $Y_i = 0$ as shown in Fig.~\ref{fig:task_edge_gnn}. Also by dropping the time index $t$, analysis in Sect.~\ref{sect:edge} will give us the Transparent Models of the two models to be BN $\mathcal{B}_1$ and $\mathcal{B}_2$ shown in Fig.~\ref{fig:dbn_edge_gnn}, respectively. WIth the same arguments as in Lemma~\ref{lemma:inexplanability_edge} we can establish the Unidentifiable Proof for Edge-perturbations of the base GNNs.

\end{document}